\documentclass{article}

\usepackage{arxiv}
\usepackage[utf8]{inputenc}
\usepackage[T1]{fontenc}
\usepackage[numbers,sort&compress]{natbib}
\usepackage[hidelinks]{hyperref}
\hypersetup{
    colorlinks=true,
    linkcolor=blue,
    citecolor=blue,
    urlcolor=blue
}
\usepackage{url}
\usepackage{microtype}
\usepackage{amsfonts}
\usepackage{float}
\usepackage{graphicx}
\usepackage{capt-of}
\usepackage{svg}
\usepackage{algorithm}
\usepackage{algpseudocode}
\usepackage{multirow}
\usepackage{booktabs}
\usepackage{cuted}
\usepackage{caption}
\usepackage{amssymb}
\usepackage{amsmath}
\usepackage{soul}
\usepackage{bm}

\newcommand{\bpmcell}[2]{\shortstack[c]{\textbf{#1}\\[1pt]{\scriptsize\bfseries$\bm{\pm}$#2}}}
\newcommand{\upmcell}[2]{\shortstack[c]{\underline{#1}\\[1pt]{\scriptsize\underline{$\pm$#2}}}}
\newcommand{\methodcell}[1]{\shortstack[c]{#1\\[1pt]{\scriptsize\vphantom{$\pm$0.0000}}}}
\newcommand{\dashcell}{\shortstack[c]{-\\[1pt]{\scriptsize\vphantom{$\pm$0.0000}}}}

\title{Toward accurate RUL and SoH estimation using reinforced graph-based physics-informed neural networks enhanced with dynamic weights}

\author{
 Mohamadreza Akbari Pour \\
  Department of Mechanical Engineering\\
  Sharif University of Technology\\
  Teymouri Square, Tarasht, Tehran, Iran\\
  \texttt{mohamadreza.akbari83@sharif.edu} \\
  \And
 Ali Ghasemzadeh \\
  Department of Computer Engineering\\
  Sharif University of Technology\\
  P.O. Box 11155-9517, Tehran, Iran\\
  \texttt{ali.ghasem01@sharif.edu} \\
  \And
 Mohamad Ali Bijarchi \\
  Department of Mechanical Engineering\\
  Sharif University of Technology\\
  Teymouri Square, Tarasht, Tehran, Iran\\
  \texttt{bijarchi@sharif.edu} \\
  \And
 Mohammad Behshad Shafii \\
  Department of Mechanical Engineering\\
  Sharif University of Technology\\
  Teymouri Square, Tarasht, Tehran, Iran\\
  \texttt{behshad@sharif.edu} \\
}

\begin{document}
\maketitle

\begin{abstract}
Accurate estimation of Remaining Useful Life (RUL) and State of Health (SoH) is essential for reliable Prognostics and Health Management (PHM), supporting timely maintenance and dependable industrial operation. However, hybrid models that combine data-driven learning with physics-based regularization often rely on fixed loss weights and therefore lose accuracy when transferred across assets with different degradation behaviors. This study introduces Reinforced Graph based Physics-informed Networks with Dynamic Weighting (RGPD), a unified framework for spatio-temporal degradation modeling and adaptive physics guided regularization. Graph-based representation learning captures inter-sensor degradation structure, a Soft Actor-Critic (SAC) module refines latent features under noisy conditions, and a lightweight Q-learning policy adaptively balances monotonicity, smoothness, and latent-dynamics residual losses during training. The framework is evaluated on the C MAPSS, PHM2012, and XJTU datasets that represent engine, bearing, and battery degradation processes. Relative to the strongest compared baselines reported in the corresponding benchmark tables, RGPD improves average RMSE by up to 12\% on PHM2012 and C MAPSS, and reduces average MAPE by 20\% on XJTU compared with the second-best reported model. Performance on these heterogeneous benchmarks further suggests the model's generalizability across degradation systems. The physics-informed component is implemented through degradation-consistent priors together with a Deep Hidden Physics Model-style residual, which improves physical plausibility without requiring a full first-principles model for each asset type.
\end{abstract}

\keywords{Remaining Useful Life, State of Health, Physics-Informed Neural Networks, Graph Neural Networks, Reinforcement Learning}
\section{Introduction}

Prognostics and Health Management (PHM) has become a cornerstone of efficiency, cost reduction, and operational reliability in today's rapidly evolving industrial landscape \citep{REN2026103925}. 
Remaining Useful Life (RUL) and State of Health (SoH) are two crucial metrics in this field that are essential for maximizing asset performance, reducing downtime, and improving safety \citep{LEI2018799}. RUL is the time before a machine or component breaks down or needs 
maintenance, whereas  SoH reflects the current condition of the system relative to its optimal or original state \citep{DING2026104069}.
The accurate estimation of RUL and SoH in engineered systems stands as a central objective within PHM applications.
A persistent challenge, however, is that degradation must be inferred from noisy, multi-sensor, and asset-dependent signals, while the resulting predictions must remain physically plausible enough to support maintenance decisions.

PHM frameworks are designed to predict the deterioration and eventual failure of a wide range of complex machinery and equipment by utilizing data-driven methodologies and cutting-edge analytical techniques. This capability holds immense practical value across numerous industrial domains, facilitates a shift toward proactive maintenance practices, and enhances system reliability \citep{SHANG202430}. Consequently, it contributes significantly to financial and operational improvements. Prediction efforts primarily fall into three main categories: physical model-based methods, which depend on knowledge of underlying degradation mechanisms; data-driven methods, which leverage historical data to develop predictive models; and hybrid methods, which integrate the strengths of both approaches \citep{LI2024111120}.
Among these, physical models rely on a deep understanding of degradation mechanisms and domain-specific knowledge. Methods such as particle filters \citep{LI2023110713} and Bayesian filtering \citep{KHINE2025115371} offer interpretability and robustness but require explicit modeling of complex systems, making them difficult to generalize across diverse applications. This creates an important trade-off: physical models provide interpretability but are often asset-specific, whereas data-driven methods are more flexible but may lack explicit degradation consistency.

Recent advances in data-driven PHM leverage deep learning for RUL estimation. Sequence-based models such as Long Short-Term Memory (LSTM) networks \citep{LI2019104785}, Gated Recurrent Units (GRUs) \citep{LIN2023110419}, and Temporal Convolutional Networks (TCNs) \citep{XU2024110288} have improved predictive performance but often struggle with rigid architectures, limited feature weighting, and insufficient modeling of multi-sensor interactions. To address these challenges, Zhang et al. \citep{ZHANG2024107241} introduced an enhanced TCN combined with Bidirectional-LSTMs, leveraging the strengths of both temporal convolution and sequential modeling, and demonstrated superior accuracy on benchmark datasets. Beyond these temporal models, Graph Neural Networks (GNNs) \citep{WANG2025112449} extend the data-driven paradigm by representing input channels as nodes and their relationships as edges, which enabled the extraction of complex spatio-temporal dependencies. Building on this, the Dual-View Graph Transformer (DVGTformer) \citep{WANG2024110935} fuses temporal and spatial dependencies via learnable adjacency and self-attention, achieving over 80\% mean relative accuracy on C-MAPSS turbofan engine data \citep{4711414}. Likewise, the Dual-Stream Spatio-Temporal Fusion Network (DS-STFN) extracts temporal and spatial features in parallel, reducing RMSE by up to 10.86\% compared to existing methods \citep{ZHANG202443}. These studies show that graph-based and spatio-temporal architectures can better exploit multi-sensor degradation structure, but they remain primarily data-driven and do not by themselves enforce physically consistent prediction trajectories.

Deep Reinforcement Learning (DRL) has emerged as a powerful tool for enabling adaptive, threshold-free maintenance by scheduling actions based on probabilistic RUL prognostics \citep{ZHANG2025127034}. While it has shown considerable promise in applications such as turbofan engines and multi-component systems, its adoption in direct RUL and SoH estimation remains limited, which highlights an opportunity to bridge predictive and prescriptive PHM strategies. For example, a recent study applied a Soft Actor-Critic (SAC)-based DRL framework that integrates autoencoder-derived features, error-aware rewards, and dynamic training termination, achieving superior accuracy on C-MAPSS and XJTU-SY datasets \citep{DING2025111121}. From the perspective of the present work, the key value of reinforcement learning is its ability to introduce adaptive control into the learning process, particularly when feature relevance or loss contributions vary during training.

These advancements have demonstrated notable improvements across diverse applications. In rolling bearings, Wang et al.~\citep{WANG2025109853} enhanced long-sequence feature extraction, leading to more accurate RUL prediction compared with conventional baselines. For turbofan engines, a hybrid Transformer GRU model reduced RMSE by 10-15\% relative to LSTM models \citep{app15105369}. Similarly, attention-based LSTMs outperformed CNN GRU baselines by leveraging 3D attention to better capture sensor correlations \citep{ 10190349}. Extending beyond mechanical systems, Yao et al.~\citep{YAO2023106437} introduced temporal spatial health indicators on SoH estimation, which achieved an RMSE as low as 0.2\% on the NASA battery dataset, demonstrating strong robustness even under limited data conditions. Nevertheless, these application-specific gains also highlight the broader challenge: high predictive accuracy in one asset type does not automatically guarantee physically reliable behavior under different degradation mechanisms, sensing structures, or operating conditions.

Despite these gains across the main applications of RUL and SoH estimation, most deep models remain purely data-driven: they do not enforce physical consistency and can be sensitive to noise and distribution shifts. In safety-critical PHM, this motivates augmenting learning with explicit physics guidance to regularize predictions and improve reliability.
To address these shortcomings, Physics-Informed Neural Networks (PINNs) have been proposed to integrate physical laws into deep learning frameworks. By embedding domain equations into the training loss, PINNs aim to enhance prediction reliability and interpretability. Recent variants, such as Liao et al.~\citep{LIAO2023102195}, incorporate attention mechanisms together with system knowledge and physics-based constraints to enhance RUL prediction accuracy in turbofan engines; however, this approach did not effectively outperform existing models and, in most subsets, remained below them in terms of prediction accuracy. PINNs have also been used for SoH prediction in lithium-ion batteries \citep{LIN2025261}, where a two-stage physics-informed neural network (TSPINN) achieved high adaptability across different chemistries and operating conditions, reducing estimation errors to as low as 0.675\% and outperforming methods based on equivalent circuit modeling and incremental capacity analysis. These hybrid models achieve improved generalization and reduce parameter count, but remain constrained by their use of static loss weights and fixed physical guidance, which can lead to instability and suboptimal convergence. Thus, the remaining issue is not simply whether to add physics, but how to balance physics-guided constraints adaptively when degradation patterns, noise levels, and asset characteristics vary.

Taken together, the literature reveals one central tension: prognostic models must learn expressive representations from heterogeneous multi-sensor data while preserving physically plausible degradation behavior. Graph-based models address the representation side of this problem by capturing sensor interactions, whereas PINNs address the consistency side by regularizing predictions with degradation-related constraints. However, existing approaches often treat these components separately or rely on fixed physics-loss weights, limiting their ability to adapt across assets with different degradation dynamics \citep{CofreMartel2021_RUL_PDE_DL, LIAO2023102195}.
To address this gap, we develop the Reinforced Graph-based Physics-informed neural networks enhanced with Dynamic weights (RGPD) framework, where graph-based spatio-temporal learning captures degradation interactions, SAC adaptively modulates hidden representations before temporal attention, and Q-learning updates physics-loss weights online through a small discrete action space. This integration reduces manual tuning and improves optimization stability while retaining physically guided learning. In this formulation, the framework does not treat representation learning, feature modulation, and physics-informed regularization as isolated components. Instead, these elements are coupled so that sensor interactions, latent feature reliability, temporal degradation patterns, and physically guided constraints are learned within the same optimization process. This formulation also reflects a broader artifact-centered modeling principle: complex engineered assets can be represented through interacting sensing channels, evolving state trajectories, and domain constraints that jointly support health-state inference.

We summarize our core contributions and the paper structure as follows. First, we present graph-based spatio-temporal representation learning that combines GATConv/GCRN with Temporal Attention Unit (TAU) to model multi-sensor degradation dependencies. Second, we introduce reinforcement-learning-based adaptive optimization, where SAC suppresses noisy latent features and Q-learning dynamically balances PINN constraint weights, reducing manual hyperparameter search. Third, we validate the framework across heterogeneous PHM scenarios, including engines, bearings, and batteries. This evaluation is not intended only as a multi-benchmark performance comparison, but also as a test of whether a unified framework can remain effective across substantially different degradation systems. We further include missing-data robustness analyses and multi-run statistics to assess stability. The remainder of the paper is organized as follows: Section~2 presents methodology and design rationale; Section~3 details datasets and experimental protocol; Section~4 reports results and ablation studies; and Section~5 discusses limitations and directions for future work.

\begin{figure}[H]           
  \centering
  \includegraphics[
    width=\textwidth,   
  ]{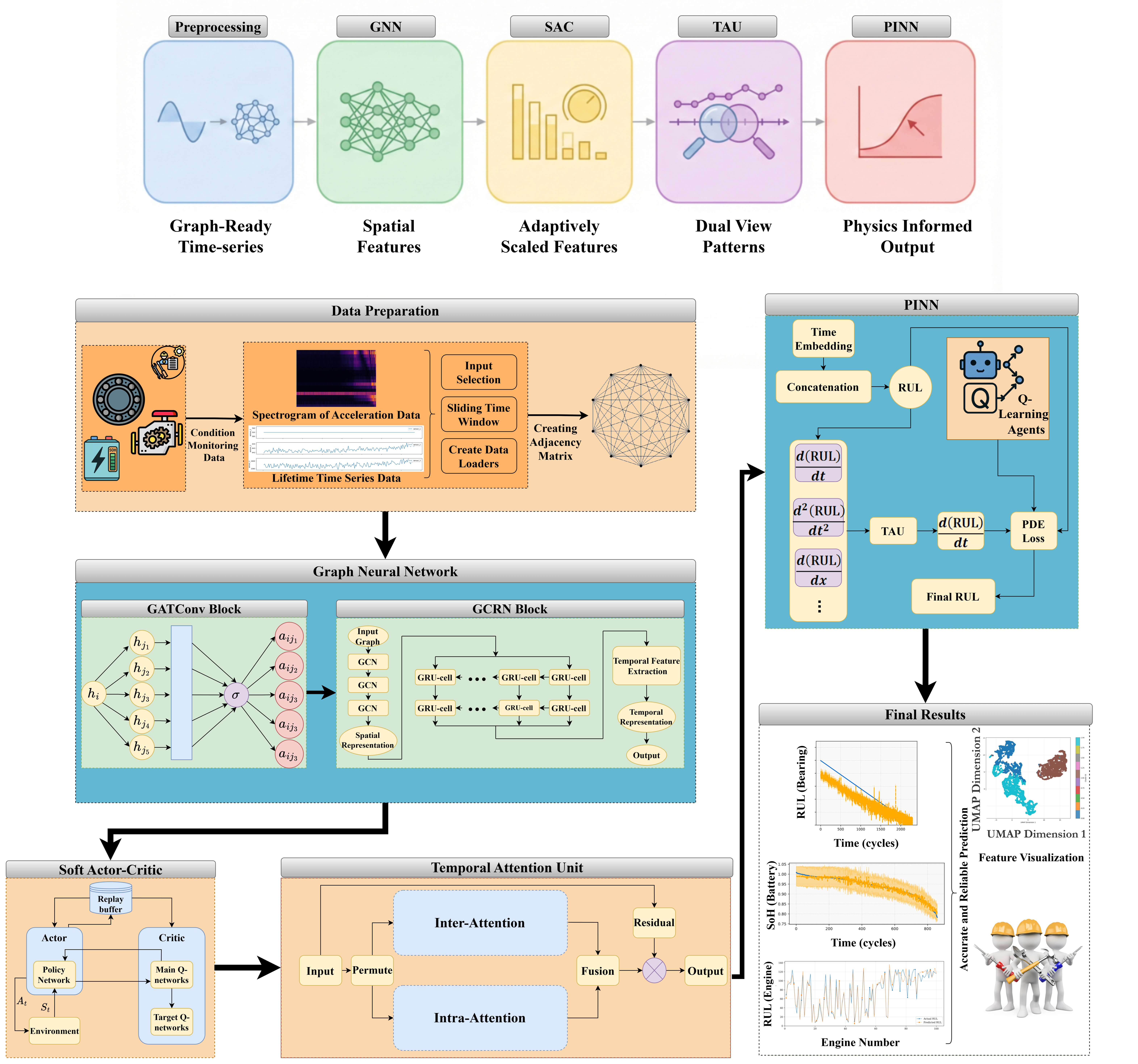}
  \caption{Overall framework of the proposed method.}
  \label{fig:overall}  
\end{figure}

An overview of the framework is sketched in \autoref{fig:overall}.

\section{Methodology}
The proposed framework is designed in a modular and problem-driven manner, where each component addresses a specific limitation in industrial prognostics. By linking representation learning, adaptive feature modulation, temporal refinement, and physics-guided regularization in a single pipeline, the framework handles noisy multi-sensor degradation signals without requiring a fixed physical model for each asset type. The GAT-GCRN backbone captures multi-sensor dependencies, SAC stabilizes latent-feature scaling under noisy training dynamics, TAU emphasizes informative temporal contexts, and PINN constraints enforce physically plausible degradation trends. Q-learning then adapts the relative weights of physics terms online to avoid static loss trade-offs across heterogeneous datasets. In this sequence, raw multi-sensor observations are progressively transformed into relational features, reliability-adjusted latent representations, temporally refined degradation patterns, and finally physics-regularized predictions. This structured integration improves accuracy, robustness, and generalization. In this way, the model links data-driven evidence, encoded degradation knowledge, and adaptive optimization in a form that can support more reliable maintenance-oriented decision-making. More generally, the framework treats prognostics as a structured transformation from artifact-level observations to decision-relevant health-state evidence.

The model input consists of windowed multi-sensor sequences, normalized time indices, and dataset-specific temporal graph edges. In tensor form, each batch is represented as \(X \in \mathbb{R}^{B \times T \times F}\), where \(B\) is the batch size, \(T\) is the window length, and \(F\) is the number of retained channels after preprocessing. The GAT-GCRN stack maps these inputs to latent spatio-temporal features in \(\mathbb{R}^{B \times T \times H}\), with \(H=64\) in the reported experiments, where \(H\) denotes the hidden feature dimension. SAC then outputs a scaling vector in \(\mathbb{R}^{B \times H}\) that rescales the latent features before TAU, TAU returns refined temporal representations in the same latent space, and the prediction head produces a degradation trajectory in \(\mathbb{R}^{B \times T \times 1}\), from which the last step is taken as the final RUL/SoH estimate. This ordering first extracts relational degradation features, then adjusts their latent reliability before temporal refinement, and finally regularizes the predicted trajectory through physics-informed losses. This keeps a consistent window-level temporal flow from graph input to trajectory output. During training, this estimate is jointly optimized with physics-informed residual losses.

\subsection{Graph Neural Networks}

GNNs have shown strong potential in time series prediction due to their ability to model complex relational dependencies among multiple interacting components. Unlike traditional sequential models that process data independently, GNNs exploit the underlying graph structure to capture both spatial correlations and their temporal evolution. This makes them particularly effective for dynamic systems where the state of each variable is influenced by its connected counterparts. In multi-sensor prognostics, this relational view helps avoid treating degradation as a set of independent temporal channels.

The proposed model integrates a GATConv layer with a GCRN to jointly capture spatial and temporal dependencies in graph-structured sequential data \citep{seo2016structuredsequencemodelinggraph}. The GATConv layer functions as a spatial encoder, assigning adaptive attention weights to neighboring nodes to prioritize the most informative relationships and suppress noisy or irrelevant ones. This selective aggregation produces context-aware node representations that better reflect the varying influence of connected entities \citep{velickovic2018graphattentionnetworks}.
These attention-refined features are then processed by the GCRN, which embeds graph convolution operations within recurrent units to model temporal dynamics conditioned on spatial dependencies. This design enables the network to learn how spatial relationships evolve over time, ensuring coherent representation of dynamic system behavior. Placing GATConv before GCRN lets the recurrent component propagate attention-refined sensor interactions rather than uniformly weighted relational signals. By feeding the GCRN with attention-enhanced inputs, the model achieves more stable temporal learning and improved representation fidelity.

To ensure reproducibility, we summarize the preprocessing and graph-construction pipeline used in code. A sliding-window representation is used in all datasets, and each window is converted to a temporal graph. For each window, each time index is treated as a graph node, and its multi-sensor vector is used as node attributes; no handcrafted edge attributes are injected, and edge importance is learned by GAT attention. For C-MAPSS, channels are Min-Max normalized after removing operating settings and selecting sensors using correlation and monotonicity thresholds. For XJTU, a 3$\sigma$ outlier filter is applied, capacity is normalized by nominal capacity, and non-capacity channels are min-max scaled. For PHM2012, the same preprocessed feature pipeline is used. Adjacency is dataset-specific: C-MAPSS uses a bidirectional temporal chain, while XJTU and PHM2012 use fully connected directed temporal graphs over window positions.

Overall, the GAT-GCRN architecture provides an efficient and adaptive mechanism for spatio-temporal representation learning. The GATConv emphasizes relevant structural information, while the GCRN captures its temporal evolution, resulting in a robust and interpretable model for complex graph-based time series prediction.
Further details on Graph Neural Networks are provided in ~\ref{appendix:GATGCRN}.

\subsection{Soft Actor-Critic}

Soft Actor-Critic is employed in our framework as an adaptive optimization mechanism to regulate feature scaling between the GCRN output and the TAU block. At this stage, the latent degradation features can contain unevenly informative or noisy dimensions, particularly under changing sensor behavior and heterogeneous operating regimes. Unlike deterministic reinforcement learning algorithms that may overfit or produce unstable gradients, SAC introduces entropy regularization into its policy objective, enabling smoother exploration and improved stability under noisy or non-stationary conditions \citep{haarnoja2018softactorcriticoffpolicymaximum}. This property makes it particularly suitable for degradation modeling, where feature distributions evolve dynamically and exhibit stochastic variability. In contrast, learned gating, channel attention, feature-wise modulation, and adaptive normalization typically rely on deterministic input-to-scale mappings optimized only through supervised loss, which can be effective in stationary settings but are less suited to sequential degradation processes where the usefulness of feature amplification or suppression depends on delayed performance feedback. Compared with these alternatives, SAC receives reward feedback and learns a stochastic control policy for feature scaling, making the modulation mechanism more adaptive to non-stationary degradation signals.

In our formulation, the SAC agent operates in a continuous control setting defined by the latent representations produced by the GCRN. The state space corresponds to these latent feature vectors, which encode both spatial and temporal dependencies of system degradation dynamics. The action space consists of continuous scaling coefficients that modulate the magnitude of latent features before they are passed to the TAU block, that allows the policy to emphasize informative dimensions and suppress noisy or redundant ones. The reward function \( r_t \) is defined as the negative supervised prediction loss,
\[
r_t = -\mathcal{L}_{\mathrm{sup}} = -\text{MSE}(y_{\mathrm{pred}}, y_{\mathrm{true}}),
\]
encouraging the policy to generate scaling patterns that minimize predictive error.

As summarized in Algorithm \autoref{alg:train_clean}, the SAC agent interacts with the model through a continuous state action reward formulation.
The SAC agent receives the latent feature map from the GCRN as the state input \( s_t \) and outputs a vector of continuous scaling coefficients \( a_t \). These coefficients modulate the latent features before they are passed to the TAU block, functioning as a dynamic normalization mechanism. The reward signal \( r_t = -\text{MSE}(y_{\text{pred}}, y_{\text{true}}) \) drives the policy to minimize prediction error by adaptively adjusting feature magnitudes. This formulation stabilizes gradient flow, prevents saturation in the TAU attention and convolution layers, and enhances convergence efficiency under stochastic training conditions. 
A detailed formulation and the associated mathematical derivations of Soft Actor-Critic can be found in ~\ref{appendix:SAC}.

\subsection{Temporal Attention Unit}

Following the SAC-based feature scaling, the processed hidden-state features are fed into the TAU, which is designed to extract temporal features while emphasizing the most relevant time steps. Because the incoming representations have already been adaptively rescaled, TAU receives a more stable feature space for identifying informative temporal contexts. By first adaptively normalizing and modulating feature magnitudes, SAC ensures that TAU receives inputs within an optimal range, preventing gradient instability and saturation in its attention and convolution operations. 

TAU leverages both convolutional and attention mechanisms \citep{10438409} to capture complex spatio-temporal patterns. Unlike conventional self-attention, which treats the temporal dimension uniformly, TAU explicitly disentangles static (intra-time) and dynamic (inter-time) dependencies through its dual-attention design. This separation enables it to capture both instantaneous correlations among features and evolving degradation trends over time, which are often missed by standard attention modules. As a result, TAU exhibits stronger temporal coherence, higher robustness to noise, and improved modeling of long-term dependencies in industrial degradation sequences \citep{10203281}.

\begin{figure}[H]           
  \centering
  \includegraphics[
    width=0.9\textwidth,   
  ]{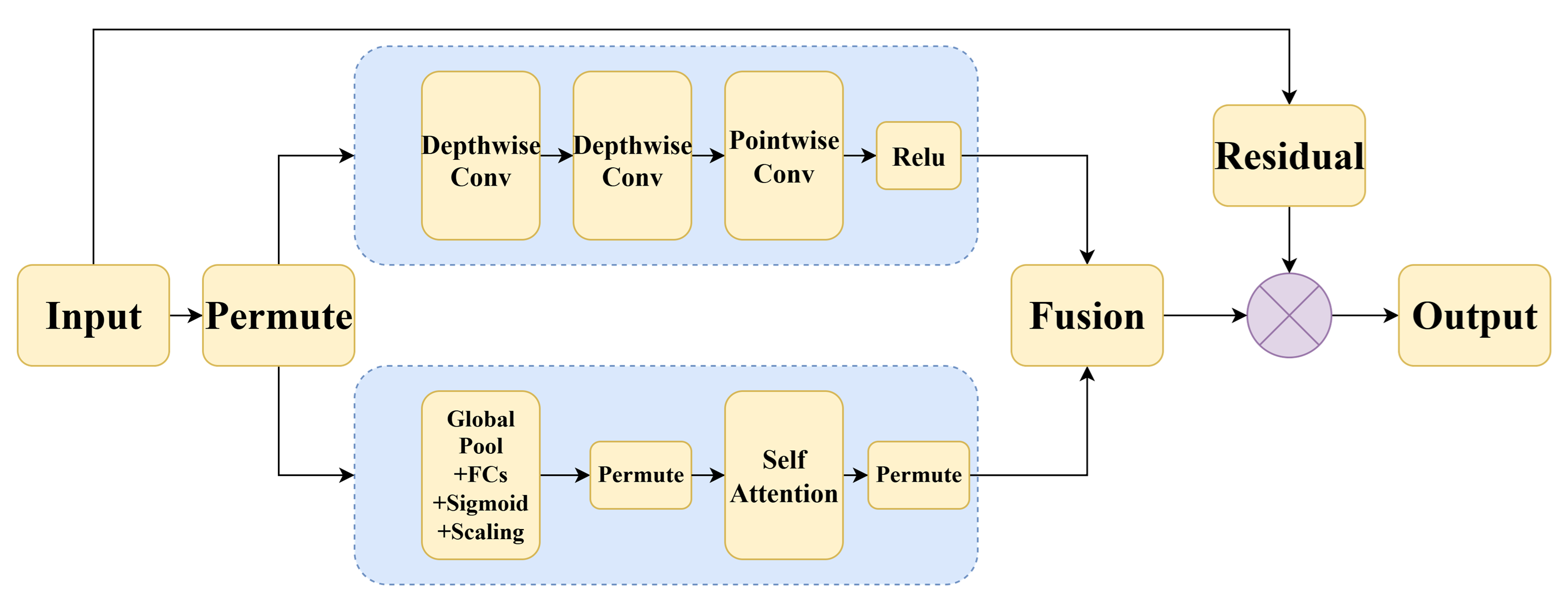} 
  \caption{Temporal attention unit.}
  \label{fig:TAU}
\end{figure}

It consists of Dynamic Attention (\(DA\)), implemented via average pooling and a fully connected layer, and Static Attention (\(SA\)), which employs Depthwise Convolutions (\(DW\)), Dilated Convolutions, and 1$\times$1 convolutions. This combination allows TAU to model long-range dependencies effectively, enhance robustness, and focus on physically meaningful temporal patterns that SAC has highlighted.

\[
SA = \mathrm{Conv}_{1\times 1}\!\big(DW\!-\!D\,\mathrm{Conv}(DW\,\mathrm{Conv}(H))\big)
\]
\[
DA = \mathrm{FC}(\mathrm{AvgPool}(H))
\]
\[
H' = (SA \otimes DA) \odot H
\]
Where $H$ is the hidden feature given to TAU, \(\otimes\) denotes the Kronecker product, and \(\odot\) denotes the Hadamard product. \autoref{fig:TAU}
represents TAU's detailed features. 
~\ref{appendix:TAU} presents a detailed description of the TAU and its constituent modules.

\subsection{PINN}

Deep Hidden Physics Models (DeepHPM) \citep{raissi2018deephiddenphysicsmodels} aim to identify the underlying partial differential equations (PDEs) that govern complex dynamical systems directly from observed data. The framework employs deep neural networks (DNNs) to jointly learn both the system solution and the structure of its dynamics. In the context of RUL prediction, this formulation was originally proposed by \citep{CofreMartel2021_RUL_PDE_DL}, who introduced a PDE-based DeepHPM framework to model degradation dynamics.

DeepHPM consists of two main components: a solution network, which approximates the unknown state function \( u(x,t) \), and a dynamics network, which represents the nonlinear operator \( \mathcal{N} \) in the governing PDE, typically expressed as
\[
u_t = \mathcal{N}(t, x, u, u_x, u_{xx}, \ldots).
\]
The model defines a residual function
\[
\mathit{pde\_residual} := u_t - \mathcal{N}(t, x, u, u_x, u_{xx}, \ldots),
\]
which ideally vanishes for a physically consistent solution. During training, this residual acts as a physics-based regularization term that constrains the neural approximation to satisfy the governing law. The temporal and spatial derivatives of \( u \) are computed analytically through automatic differentiation, allowing exact evaluation of \( u_t \), \( u_x \), and higher-order terms without numerical discretization. This enables DeepHPM to learn a smooth, continuous, and physically consistent representation of the system dynamics.
In the present model, the DeepHPM formulation is employed solely as one component of the overall physical constraints, rather than as a complete physics-based modeling framework. This keeps the physics-informed component compatible with heterogeneous assets, where detailed first-principles models are not always available.

To clarify the role of the different regularization terms in the model, we distinguish between two complementary levels of physical guidance. The first level is the DeepHPM residual, which captures latent continuous-time degradation dynamics through a learned PDE structure. The second level consists of three auxiliary trajectory-level constraints, namely monotonicity, smoothness, and broken-device enforcement. These auxiliary terms are not intended to describe the exact component-level physics of every system; rather, they encode general prognostic priors that are meaningful when the output is defined as a normalized SoH or RUL trajectory. In this setting, engines, bearings, and lithium-ion batteries share common degradation-level behaviors: health indicators are expected to deteriorate with usage, degradation trajectories should not exhibit unrealistic abrupt oscillations, and the predicted RUL should approach zero at the end-of-life condition.

Accordingly, monotonicity and smoothness are used as general degradation priors, while the broken-device constraint represents an engineering boundary condition. The DeepHPM residual complements these terms by learning the hidden degradation dynamics from data. Similar PINN-based prognostics studies have used DeepHPM-type residuals or partial forms of such degradation-aware constraints \citep{Lu2023, LIAO2023102195, wang2024physics, CofreMartel2021_RUL_PDE_DL}; however, their joint use with monotonicity, smoothness, and broken-device losses within a unified adaptively weighted objective has not been explicitly addressed in these prior works. Together with the DeepHPM residual, these terms provide complementary forms of physical guidance, combining target-level degradation behavior with learned latent dynamics.

\subsubsection{Physical Constraints}

The model enforces basic but informative physical constraints on SoH and RUL estimation, denoted by $\hat{y}_t$, to ensure consistent degradation dynamics, monotonicity, smoothness, and physical realism. These constraints are integral to the model's design and are coupled with the PINN framework to ensure reliable predictions. The constraints are implemented alongside the neural PDE loss, guiding the model to obey physical laws during training. Because these constraints act on the predicted trajectory, they remain applicable even when the detailed component-level physics differs across asset types.

The monotonicity constraint ensures that the RUL and SoH metrics decrease over time, reflecting the natural degradation of the system. Mathematically, this is enforced by the condition $\frac{d\hat{y}}{dt} \leq 0$, which ensures that the predicted values of $\hat{y}_t$ do not increase. The corresponding difference and residual are computed as:

\[
\mathit{monotonicity_diff} = \hat{y}_{t+1} - \hat{y}_t, \quad \mathit{monotonicity_residual} = \max(0, \mathit{monotonicity_diff})
\]

Next, the smoothness constraint is applied to ensure that the rate of degradation changes smoothly over time, avoiding abrupt shifts. The smoothness is enforced by:

\[
\mathit{smoothness_diff} = (\hat{y}*{t+2} - \hat{y}*{t+1}) - (\hat{y}_{t+1} - \hat{y}_t)
\]

Finally, a failed-device constraint is included to ensure that the model predicts a zero RUL when a device has failed or when SoH falls below a threshold of 0.8. This constraint prevents unrealistic predictions for failed systems. The broken loss is calculated as:

\[
\mathit{broken_loss} = \operatorname{mean}(\mathit{broken_mask} \cdot \hat{y}_{\mathit{last}}^2)
\]

This loss is then added to the PDE loss, ensuring that the system behaves correctly when the device is considered failed. In the final objective, the DeepHPM residual and the three auxiliary losses are jointly optimized, with their relative contributions adaptively adjusted by Q-learning. The weighting process is therefore linked to validation behavior rather than treated as a fixed hyperparameter choice. Each physics-related term contributes a residual-based state, and the selected action determines how strongly that term influences the next stage of optimization.

\subsubsection{Q-learning Agents}

While DeepHPM provides a framework for learning the underlying physics of a dynamical system, the performance of the model depends critically on balancing multiple physics-based loss terms. To automate this process and dynamically adapt the contributions of each term, we employ Q-learning \citep{watkins1992qlearning}. This is especially relevant because fixed PINN loss weights can become mismatched as different constraints dominate at different stages of training.
Q-learning is one of the foundational algorithms in RL. It is a model-free and off-policy method that learns the action-value function $Q(s,a)$ using
\[
Q(s_t, a_t) \leftarrow Q(s_t, a_t) + \alpha [r_{t+1} + \gamma \max_{a'} Q(s_{t+1}, a') - Q(s_t, a_t)].
\]
In the current implementation, dynamic weighting is handled by four coordinated Q-learning agents, where each agent is assigned to one physics-related loss component. Agent~1 controls the weight of the first PDE-residual constraint based on its residual magnitude, Agent~2 adjusts the weight of the smoothness-related degradation difference term, Agent~3 regulates the consistency between the degradation derivative and the latent degradation variable, and Agent~4 adapts the broken-device penalty weight. Unlike a shared weighting strategy, this formulation enables each constraint to be updated according to its own training state and contribution to validation performance. Using separate agents also avoids a combinatorial joint-action space and keeps each policy aligned with the scale and dynamics of one loss term. The discrete levels are intentionally coarse and bounded around unit weight so that the controller can represent under-emphasis, nominal emphasis, and over-emphasis without a costly continuous search. As a result, the framework can adaptively emphasize or relax different physical constraints across datasets with heterogeneous degradation patterns, while maintaining a compact and interpretable discrete action space.

\begin{itemize}
    \item Agent 1 selects $w_1$ from a discrete action space $\{0.1,0.5,1.0,2.0,5.0,10.0\}$ based on state \(s_1 = \operatorname{mean}(\mathit{pde\_residual}_1^2)\).
    \item Agent 2 selects $w_2$ from $\{0.01,0.05,0.1,0.5,1.0,2.0\}$ based on state \(s_2 = \operatorname{mean}(\mathit{diff}_2^2)\).
    \item Agent 3 selects $w_3$ from $\{0.1,0.5,1.0,2.0,5.0,10.0\}$ based on state $s_3 = \operatorname{mean}\!\left((\mathit{diff}_1 - N_u)^2\right)$.
    \item Agent 4 selects $w_4$ from $\{0.1,0.5,1.0,2.0,5.0,10.0 \}$ based on state \(s_4 = \mathit{broken\_loss}\).
    \item Q-tables are updated using reward $r=10\cdot(\text{prev\_rmse}-\text{valid\_rmse})$ with $\epsilon$-greedy exploration ($\epsilon=0.1$), learning rate $0.1$, and discount factor $0.95$.
\end{itemize}

The selected actions correspond directly to the adaptive weights in the total loss. Therefore, the Q-learning agents do not replace gradient-based neural-network training; instead, they determine how strongly each physical constraint contributes to the loss used for backpropagation. The action spaces are intentionally small and bounded so that each agent explores only meaningful low/medium/high emphasis levels, which is more stable and sample-efficient than continuous weight search in small or medium PHM datasets. In addition, the chosen values span attenuation and amplification around the nominal scale, which is useful because the four physics terms do not have the same magnitude or sensitivity across datasets. At each training stage, the total loss is computed as
\[
\mathit{Total\_loss} = w_1 \cdot \operatorname{mean}(\mathit{monotonicity\_residual}^2)
+ w_2 \cdot \operatorname{mean}(\mathit{smoothness\_diff}^2)
+ w_3 \cdot \operatorname{mean}(\mathit{pde\_residual}^2)
+ w_4 \cdot \mathit{broken\_loss}.
\]
The neural model is optimized batch-wise using gradient descent, while the Q-tables are updated after validation evaluation because the reward is defined from the change in validation RMSE. Thus, actions that reduce validation RMSE receive positive feedback, whereas actions that increase validation error are penalized. The bounded discrete action spaces keep the search stable and sample-efficient, while still allowing the model to adaptively rebalance the physical constraints across heterogeneous PHM datasets.

Conceptually, the proposed strategy differs from adaptive weighting approaches that rely directly on gradient magnitudes or instantaneous loss statistics during optimization. Instead, weight selection is formulated as a validation-driven multi-agent decision process, in which each agent learns a separate policy over its residual-based state and updates the corresponding loss weight according to its impact on validation performance.

\begin{figure}[H]           
  \centering
  \includegraphics[
    width=0.7\textwidth,  
  ]{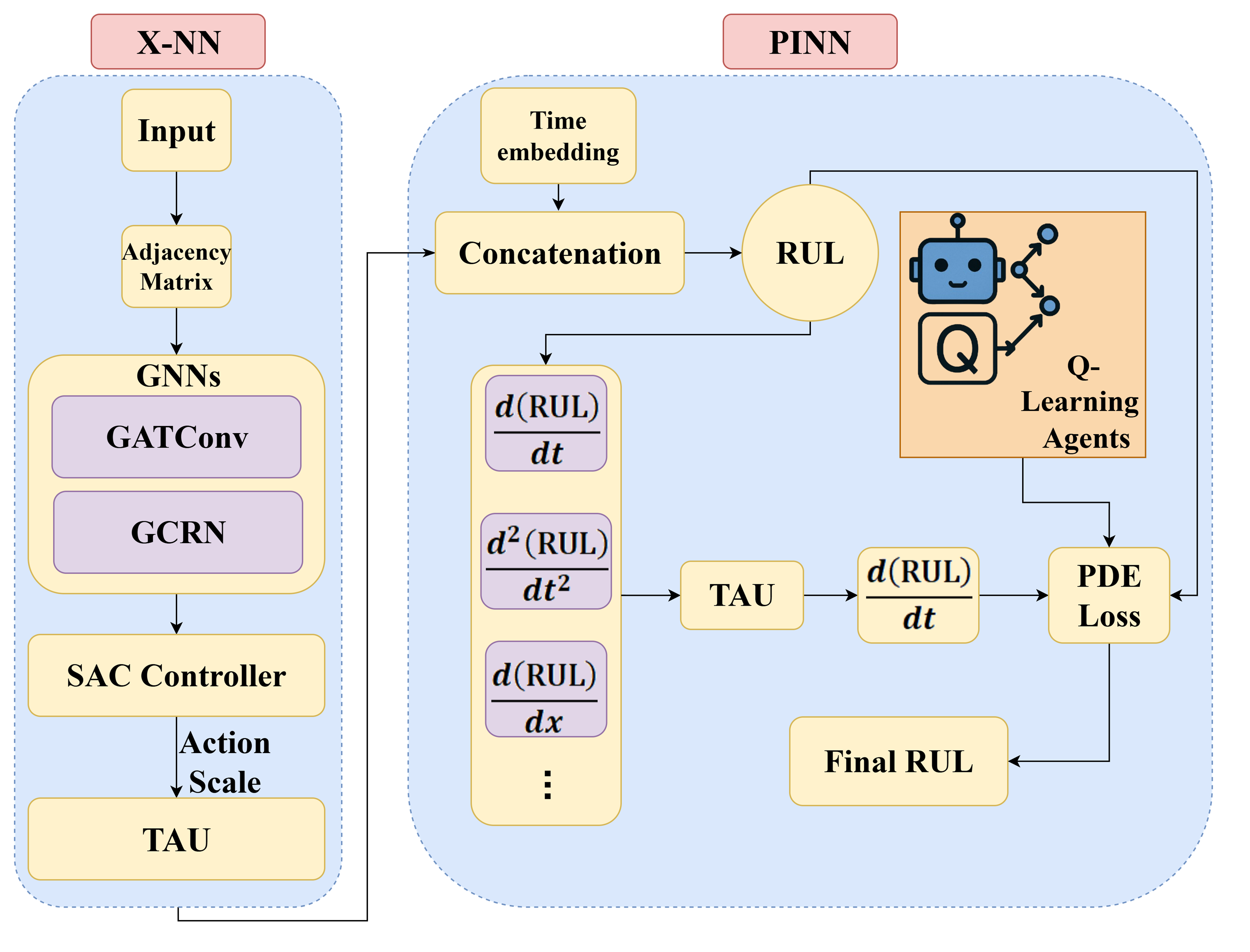} 
  \caption{Proposed PINN integrated graph-based neural network.}
  \label{fig:deephpm} 
\end{figure}

As detailed in Algorithm~\ref{alg:forward_rl}, after the reward signal is computed from validation RMSE improvement, each agent updates its Q-table using the temporal-difference rule and progressively favors weight choices that improve validation performance.
An overview of the proposed PINN structure is shown in \autoref{fig:deephpm}.

\subsection{Algorithm Summary}
The complete training pipeline is summarized below:

\begin{algorithm}[H]
  \caption{Training procedure.}
  \label{alg:train_clean}
  \begin{algorithmic}[1]
  \Require Preprocessed datasets $\{\mathcal{D}_{\text{train}}, \mathcal{D}_{\text{valid}}, \mathcal{D}_{\text{test}}\}$ and configuration hyperparameters
  \For{$i = 1$ to $D$}
    \State Build DataLoaders \texttt{train}, \texttt{val} from $\mathcal{D}_{\text{train}}, \mathcal{D}_{\text{valid}}$
    \State Construct temporal graph $\texttt{edge\_index}$ for window length $\lvert T_{\text{train}}^{(i)} \rvert$
    \State \textbf{Initialize} $M^{(i)} \leftarrow \mathrm{custom\_model}(\mathrm{config})$
    \State Initialize optimizer (Xavier-uniform init) and scheduler
    \For{each epoch}
      \For{each batch $(x,t,y)$ in \texttt{train}}
        \State $(\hat{y},\,\mathcal{L}_{\mathrm{PDE}},\,s,\,a,\,\log p)
        \leftarrow M^{(i)}(x,\,t,\,\texttt{edge\_index};\,\mathrm{targets}=y)$
        \State $\mathcal{L}_{\text{sup}} \leftarrow \mathrm{CE}(\hat{y}, y)$
        \State $\mathcal{L} \leftarrow \mathcal{L}_{\text{sup}} + w_{\mathrm{pde}} \cdot \mathcal{L}_{\mathrm{PDE}}$
        \State Backpropagate $\mathcal{L}$ and update $M^{(i)}$ every $k$ steps
        \State Push $(s, a, r = -\mathcal{L}_{\text{sup}}, s, 1.0)$ into SAC buffer
        \State Update SAC agent
      \EndFor
      \State Evaluate validation RMSE
      \State Update Q-learning agents using $r=10\cdot(\text{previous\_rmse}-\text{valid\_rmse})$
    \EndFor
  \EndFor
  \end{algorithmic}
\end{algorithm}

\begin{algorithm}[H]
    \caption{RGPD approach.}
    \label{alg:forward_rl}
    \begin{algorithmic}[1]
    \Require 
      Input features $X\in\mathbb R^{B\times T\times D}$,  
      time-stamps $t\in\mathbb R^{B\times T}$,  
      graph edges $E$,  
      true labels $Y\in\mathbb R^B$
    \State \textbf{1. Spatial--Temporal Feature Encoding}
    \State \quad $H \leftarrow$ apply GATConv on $(X, E)$ 
    \State \quad $H_G \leftarrow$ apply GCRN block on $(H, E)$
    \State \quad $s \leftarrow$ mean over $H_G$ \Comment{global state for SAC}
    \vspace{4pt}
    \State \textbf{2. SAC-Based Feature Modulation}
    \State \quad $(\text{action},\,\log p)\leftarrow$ apply SACController on $s$  
    \State \quad $H_G \leftarrow H_G \cdot (1 + \text{action})$
    \vspace{4pt}
    \State \textbf{3. TAUBlock \& Time Embedding}
    \State \quad $H_T \leftarrow$ apply TAU\_block on $H_G$
    \State \quad $\text{T}_{\mathrm{emb}} \leftarrow$ apply $\mathrm{timeEmbed}$ on $\frac{t}{T_{\max}}$
    \State \quad $F \leftarrow \mathrm{concat}(H_T,\,T_{\mathrm{emb}})$
    \vspace{4pt}
    \State \textbf{4. RUL Prediction}
    \State \quad $\hat Y_{\mathrm{seq}} \leftarrow \mathrm{Linear}(F)$
    \vspace{4pt}
    \If{training}
      \State \textbf{5. Physics--Informed Residuals}
      \State \quad $\Delta_1, \Delta_2, du\_dx, N_u \leftarrow$ apply physics roles and DeepHPM model
    
      \vspace{4pt}
      \State \textbf{6. Q-Learning for PDE Weights}
      \State \quad $w1\_state = \mathbb{E}\left[\left(\frac{\partial \hat{y}}{\partial t}\right)^2\right]$
      \State \quad $w2\_state = \mathbb{E}\left[\left(\frac{\partial^2 \hat{y}}{\partial t^2}\right)^2\right]$
      \State \quad $w3\_state = \mathbb{E}\left[\left(\frac{\partial \hat{y}}{\partial t} - \mathcal{N}_u\right)^2\right]$
      \State \quad $w4\_state = \mathbb{E}\left[\mathbb{I}(Y=0) \cdot \hat{Y}_{\mathrm{seq}}[:,-1]^2\right]$
      \State \quad Select $w_1,w_2,w_3,w_4$ using the four Q-learning agents
      \State \quad After validation, update Q-tables using $r = 10\cdot(\text{previous\_rmse}-\text{valid\_rmse})$
    
      \vspace{4pt}
      \State \textbf{7. Weighted PDE Loss \& Broken-Device Constraint}
      \State $\mathcal{L}_{\mathrm{PDE}} \leftarrow
        w_1\|\Delta_1\|^2 
        + w_2\|\Delta_2\|^2 
        + w_3\|\Delta_1 - N_u\|^2 
        + w_4\mathbf{1}_{Y=0}\,\|\hat Y_{\mathrm{seq}}[:,-1]\|^2$
    \EndIf
    \vspace{4pt}
    \State \Return $\hat{Y}_{\mathrm{seq}}[:,-1],\;\mathcal{L}_{\mathrm{PDE}},\;s,\;a,\;\log p$
    \end{algorithmic}
\end{algorithm}

\section{Experimental Validation}

The main training hyperparameters used across the reported experiments are summarized in \autoref{tab:ablation_hparams}. A sliding-window representation is used for all datasets (multi-window in C-MAPSS and fixed-window in XJTU/PHM2012). The TAU factor is a post-TAU scaling coefficient that calibrates temporal-feature amplitude before the prediction head.

\begin{table}[H]
  \centering
  \caption{Main hyperparameter settings used in experiments.}
  \label{tab:ablation_hparams}
  \renewcommand{\arraystretch}{1.2}
  \resizebox{\textwidth}{!}{
  \begin{tabular}{lccc}
    \toprule
    Hyperparameter & C-MAPSS & XJTU & PHM2012 \\
    \midrule
    Batch size & 64 & 32 & 64 \\
    Max epochs / patience & 80 / 15 & 200 / 15 & 80 / 15 \\
    Optimizer (supervised) & AdamW & AdamW & AdamW \\
    Supervised learning rate & $5\times10^{-4}$ & $10^{-3}$ & $5\times10^{-4}$ \\
    GCRN hidden size & 64 & 64 & 64 \\
    GAT heads & 4 & 2 & 4 \\
    Dropout & 0.1 & 0.1 & 0.1 \\
    TAU factor & 3.0 & 5.0 & 3.0 \\
    SAC actor / critic lr & $3\times10^{-4}$ / $3\times10^{-4}$ & $4\times10^{-5}$ / $1.21\times10^{-5}$ & $3\times10^{-4}$ / $3\times10^{-4}$ \\
    SAC $(\gamma,\tau,\alpha)$ & (0.99, 0.005, 0.2) & (0.99, 0.005, 0.2) & (0.99, 0.005, 0.2) \\
    Q-learning $(\alpha,\gamma,\epsilon)$ & (0.1, 0.95, 0.1) & (0.1, 0.95, 0.1) & (0.1, 0.95, 0.1) \\
    \bottomrule
  \end{tabular}
  }
\end{table}

\vspace{2pt}

\subsection{Results and Comparison on Different Datasets}

\subsubsection{Performance evaluation metrics}
The performance evaluation metrics used in the following tables are defined as:
\begin{flalign*}
\text{Mean Absolute Error (MAE)} &= \frac{1}{N} \sum_{i=1}^{N} \left| \hat{y}_i - y_i \right| &
\\
\text{Root Mean Squared Error (RMSE)} &= \sqrt{ \frac{1}{N} \sum_{i=1}^{N} \left( \hat{y}_i - y_i \right)^2 } &
\\
\text{Score} &= \sum_{i=1}^{N}
\begin{cases}
e^{-\frac{e_i}{13}} - 1, & \text{if } e_i < 0 \\
e^{\frac{e_i}{10}} - 1,  & \text{if } e_i \geq 0
\end{cases}, \quad e_i = y_i - \hat{y}_i &
\\
\text{Mean Absolute Percentage Error (MAPE)} &= 
\frac{100\%}{N} \sum_{i=1}^{N} \left| \frac{e_i}{y_i} \right|, 
\quad e_i = y_i - \hat{y}_i &
\end{flalign*}

To clarify baseline provenance, the comparison tables in this section report values from the cited publications unless explicitly stated as reimplemented under our internal pipeline. In practice, straightforward baselines such as MLP, CNN, and LSTM were reimplemented and evaluated under the same preprocessing, windowing, splitting, and metric protocol, while more specialized methods were taken from the original papers. In contrast, the ablation models in this paper were trained under a unified setup for module-level attribution.

\subsubsection{C-MAPSS dataset}
\begin{figure}[H]  
  \centering
  \includegraphics[width=\textwidth]{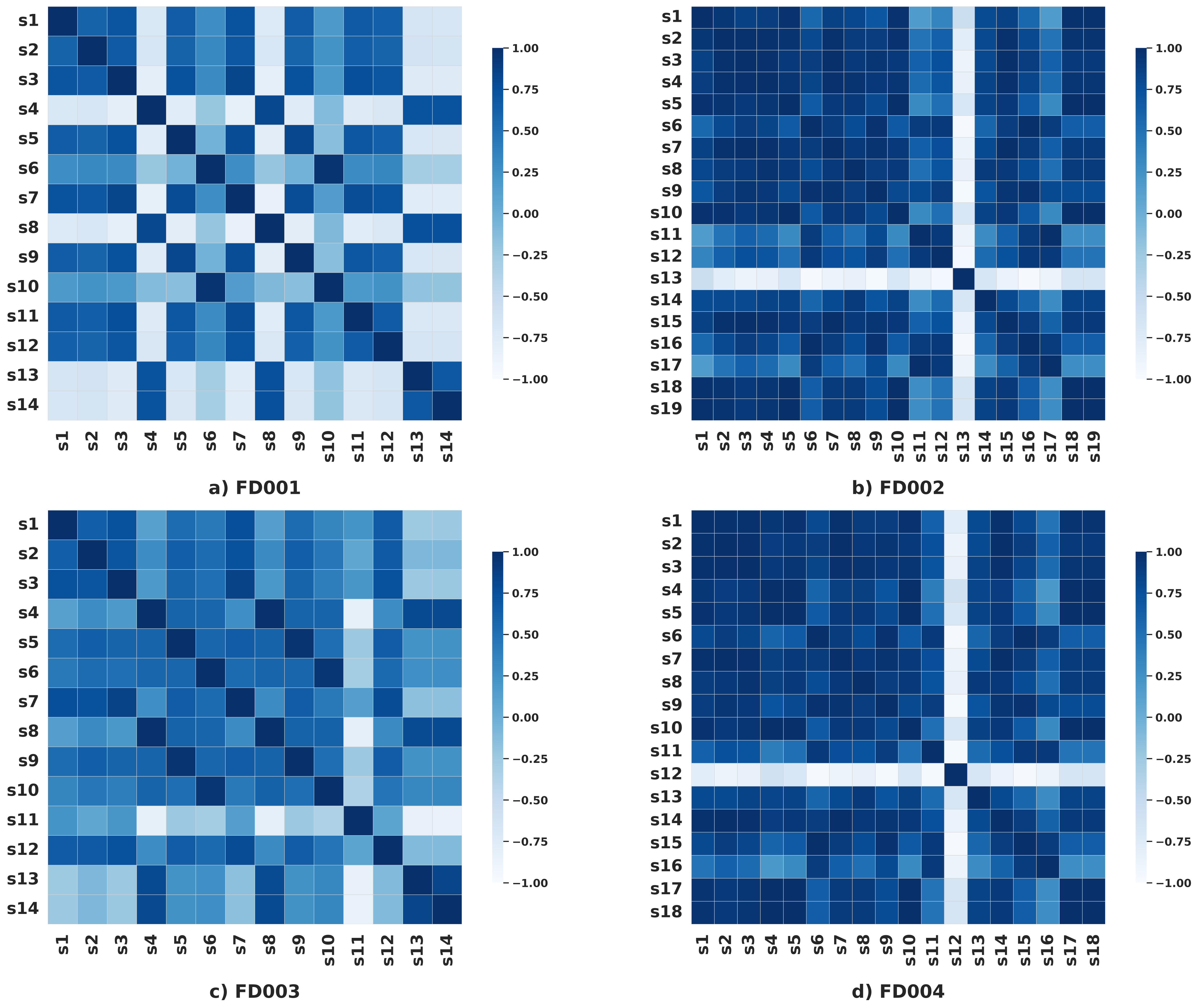}
  \caption{Correlation matrices of C-MAPSS subsets.}  \label{fig:corr}
\end{figure}

The C-MAPSS dataset serves as a simulated run-to-failure benchmark for engines. It contains 4 subsets (FD001-FD004) that differ in the number of operating conditions and fault modes, and each subset provides complete trajectory runs to failure and test trajectories truncated before failure. At each cycle, three operational settings and 21 input channels are recorded, forming a rich multivariate time series. FD001 and FD003 simulate a single degradation mode under one operating condition, while FD002 and FD004 introduce six operating scenarios and two concurrent fault modes, increasing prognostic challenge complexity.
Thanks to its synthetic yet physics-based generation and standardized structure, C-MAPSS remains a prominent benchmark dataset for RUL estimation, underpinning hundreds of machine-learning studies and fostering reproducible comparisons across prognostic models. In \autoref{fig:corr}, the correlation matrices of C-MAPSS subsets are depicted; constant or less-informative input channels are eliminated. FD002 and FD004 contain more informative input channels than FD001 and FD003, consistent with \autoref{C-MAPSS_datasets}, which shows that FD002 and FD004 involve multiple operating conditions and, for FD004, multiple fault modes, making them the most complex subsets.

\begin{table}[H]
    \centering
    \small
    \setlength{\tabcolsep}{4pt} 
    \renewcommand{\arraystretch}{1.2}
    \caption{C-MAPSS dataset: operating conditions, fault modes, and trajectory counts.}
    \label{C-MAPSS_datasets}
    \begin{tabular}{c c c c c}
        \toprule
        Dataset & Train Trajectories & Test Trajectories & Conditions & Fault Modes \\
        \midrule
        FD001 & 100 & 100 & 1 (Sea Level) & 1 (HPC Degradation) \\
        FD002 & 260 & 259 & 6 & 1 (HPC Degradation) \\
        FD003 & 100 & 100 & 1 (Sea Level) & 2 (HPC Degradation, Fan Degradation) \\
        FD004 & 248 & 249 & 6 & 2 (HPC Degradation, Fan Degradation) \\
        \bottomrule
    \end{tabular}
\end{table}

The study employed a multi-time-window approach with three distinct window sizes for each sub-dataset, consistent with the method used by Li et al.~\citep{LI2023101898} and Zhou et al.~\citep{ZHOU2022344} for data partitioning in RUL prediction of C-MAPSS dataset.

\autoref{fig:corr} 
shows that FD001 and FD003 have generally lower correlations, indicating more independent channel behaviors, while FD002 and FD004 have stronger inter-channel correlations, suggesting complex operational conditions and possible redundancy, according to the heatmaps showing the linear relationships between input channel readings across the four C-MAPSS subsets.

\begin{table}[H]
    \centering
    \caption{Results comparison for C-MAPSS.}    \label{Results_comparison_sorted_desc}
    \resizebox{\textwidth}{!}{%
    \begin{tabular}{l c c c c c c c c c c c}
        \toprule
        \multirow{2}{*}{Model} & \multirow{2}{*}{Year} 
            & \multicolumn{2}{c}{FD001} 
            & \multicolumn{2}{c}{FD002} 
            & \multicolumn{2}{c}{FD003} 
            & \multicolumn{2}{c}{FD004} 
            & \multicolumn{2}{c}{Average} \\
        \cmidrule(lr){3-4} \cmidrule(lr){5-6} 
        \cmidrule(lr){7-8} \cmidrule(lr){9-10} 
        \cmidrule(lr){11-12}
        & & RMSE & Score & RMSE & Score & RMSE & Score & RMSE & Score & RMSE & Score \\
        \midrule

        Attention-LSTM \citep{10144080}        & 2023 
            & 15.45       & 455.92     
            & 20.91       & 3602.94    
            & 14.67       & 473.97     
            & 24.01       & 6841.82    
            & 18.76       & 2843.66    \\

        ABGRU      \citep{LIN2023110419}       & 2023 
            & 12.83       & 221.54     
            & 17.97       & 2072.21    
            & 13.23       & 279.18     
            & 21.55       & 3625.77    
            & 16.39       & 1549.67    \\

        CATA-TCN   \citep{LIN2024102372}       & 2024 
            & 12.80       & 234.31     
            & 17.61       & 1361.23    
            & 13.16       & 290.63     
            & 21.04       & 2303.42    
            & 16.15       & 1047.40    \\

        BiGRU-TSAM \citep{ZHANG2022108297}     & 2022 
            & 12.60       & 213.00     
            & 18.90       & 2264.00    
            & 12.50       & 233.00     
            & 20.50       & 3610.00    
            & 16.12       & 1580.00    \\

        MmoE-BiGRU  \citep{ZHANG2022108263}     & 2022 
            & 13.22       & \multicolumn{1}{c}{N/A} 
            & 18.26       & \multicolumn{1}{c}{N/A}
            & 13.79       & \multicolumn{1}{c}{N/A}
            & 18.38       & \multicolumn{1}{c}{N/A}
            & 15.91       & \multicolumn{1}{c}{N/A} \\

        GAT-EdgePool \citep{LI2022108653}     & 2022 
            & 13.53       & 309.59     
            & 15.07       & 1380.41    
            & 14.66       & 470.74     
            & 17.60       & 1726.53    
            & 15.22       & 971.82     \\

        CNN-LSTM-SAM \citep{10089402}          & 2023 
            & 12.60       & 261.00     
            & 15.30       & 1156.00    
            & 13.80       & 253.00     
            & 18.60       & 2425.00    
            & 15.08       & 1023.75    \\

        DA-TCN     \citep{9123333}             & 2021 
            & 11.78       & 229.48     
            & 16.95       & 1842.38    
            & 11.56       & 257.11     
            & 18.23       & 2317.32    
            & 14.63       & 1161.57    \\

        MTSTAN     \citep{LI2023101898}        & 2023 
            & \underline{10.97}  & \underline{175.36}  
            & 16.81       & 1154.36    
            & \underline{10.90}  & \textbf{188.22}  
            & 18.85       & 1446.20    
            & 14.38       & 741.53     \\

        GF-GGAT    \citep{WANG2025110902}      & 2025 
            & 11.71       & 188.00  
            & 14.92       & 990.47     
            & 13.06       & 247.21  
            & 15.71       & \underline{996.66}  
            & 13.85       & 605.09  \\

        PSTFormer  \citep{FU2025125995}        & 2024 
            & 12.08       & 224.00     
            & \underline{13.00}  & 877.00  
            & 12.11       & 308.00     
            & \underline{14.38}  & 1182.00    
            & \underline{12.89}  & 647.75     \\

        PVA-FFG-Transformer \citep{ZHOU2024567} & 2024
            & 11.36 & \textbf{173.89}
            & 13.24 & \underline{714.98}
            & 11.80 & \underline{215.32}
            & 15.16 & 1049.53
            & 12.89 & \underline{538.43} \\

        AttnPINN \citep{LIAO2023102195} & 2023 
            & 16.89 & 523 
            & 16.32 & 1479 
            & 17.75 & 1194 
            & 18.37 & 2059 
            & 17.33 & 1313.75 \\

        \methodcell{\textbf{RGPD (Proposed)}}      
            & \dashcell
            & \bpmcell{10.77}{0.23}   & \upmcell{207.09}{34.1}
            & \bpmcell{11.58}{0.49}   & \bpmcell{684.24}{167.8}
            & \bpmcell{10.78}{0.16}   & \upmcell{245.28}{43.2}
            & \bpmcell{12.38}{0.57}   & \bpmcell{787.39}{249.0}
            & \bpmcell{11.38}{0.36}   & \bpmcell{481.00}{123.5} \\[-0.3ex]
        \addlinespace[0.4ex]
           
        \bottomrule
    \end{tabular}%
    }
    \begin{flushleft}
    The best results are \textbf{bolded}, and the second-best results are \underline{underlined}.\\
    \end{flushleft}
\end{table}

\begin{figure}[H]       
  \centering
  \includegraphics[
    width=0.8\textwidth,  
  ]{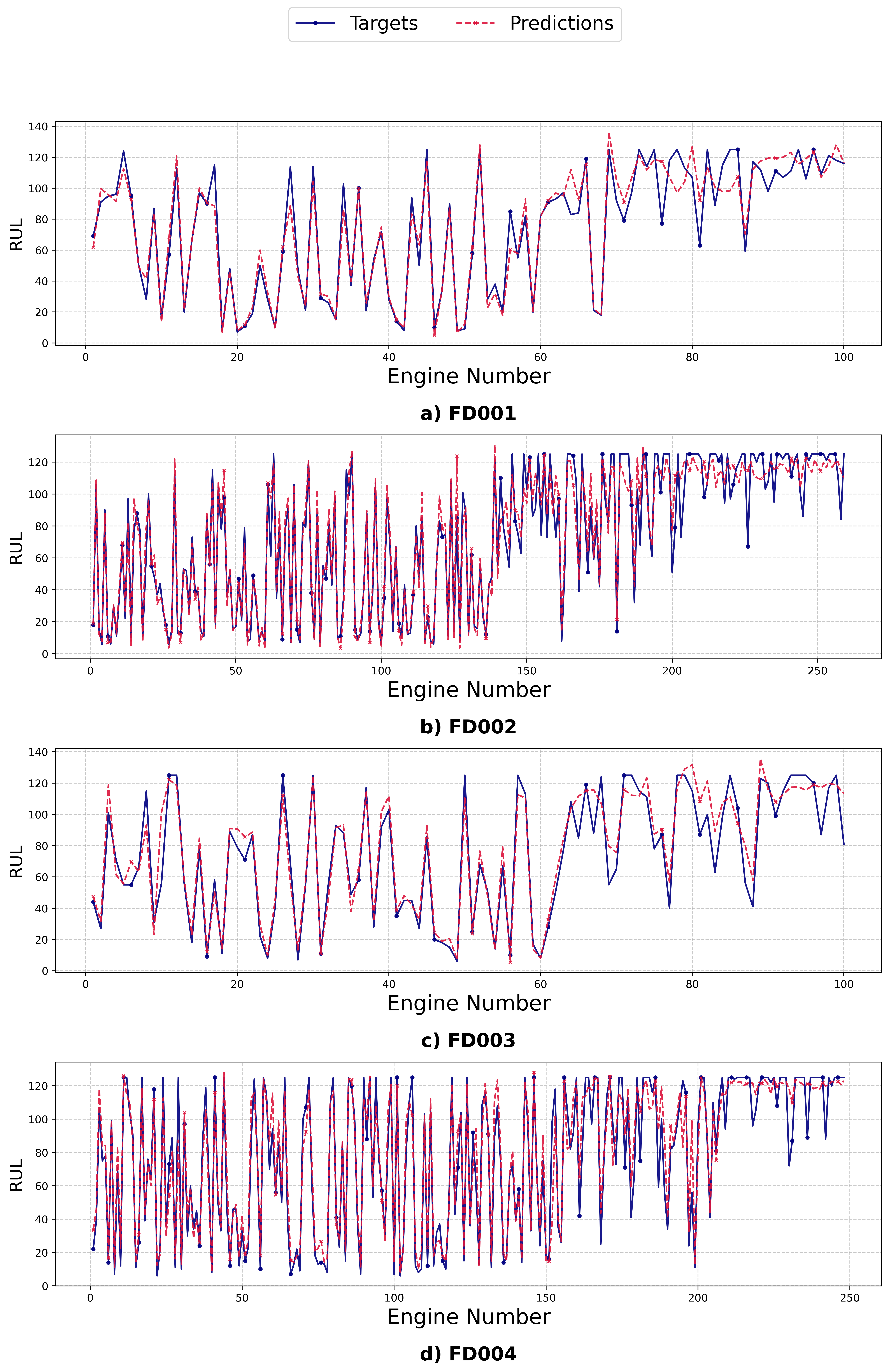}
  \caption{Predicted versus true RUL values for C-MAPSS test engines across four subsets: 
(a) FD001, (b) FD002, (c) FD003, and (d) FD004. Solid blue lines denote the true RUL, 
while red dashed lines with circle markers indicate the predictions.}

  \label{fig:C-MAPSS-sort}   
\end{figure}

In \autoref{fig:C-MAPSS-sort}, the predicted and true RUL values are compared across the four C-MAPSS subsets.
The model achieves higher accuracy in FD001 and FD003, where the datasets involve fewer engines and a limited number of fault modes, thus reducing variability. In contrast, FD002 and FD004 exhibit larger prediction errors due to the presence of multiple operating conditions and diverse fault modes, which increase complexity.
Nevertheless, the predictions still follow the overall RUL degradation trend, demonstrating the model robustness in simpler scenarios and its reasonable adaptability to more challenging conditions.

As detailed in 
\autoref{Results_comparison_sorted_desc}
, the proposed model demonstrates exceptional performance in the C-MAPSS dataset and sets a new standard for RUL prediction. It achieves superior accuracy and reliability, as evidenced by its minimized RMSE and Score values across diverse operational contexts. The model's robustness is particularly notable in complex scenarios involving multiple operating conditions and fault modes, where it maintains precise and stable predictions. This consistency of excellence in various challenges underscores its generalizability and technical sophistication.

In addition to its robust performance, the proposed model introduces several key improvements that further enhance its capabilities. In particular, it achieves a significant reduction of 11.71\% in average RMSE compared to the previous state-of-the-art, demonstrating a substantial leap in prediction accuracy. This is complemented by a 10.59\% improvement in the average score metric, indicating not only precise but also timely predictions.

\autoref{fig:rul_comparison} shows RUL predictions for four sample engines from the C-MAPSS subsets. The model closely tracks the degradation trend, with predictions aligning well with the true RUL. The confidence intervals are wider in early cycles, where degradation evidence is less distinctive and different training runs may produce more dispersed trajectories, and shrink toward end-of-life as the failure trend becomes clearer, indicating higher reliability near failure, which is the most critical stage for prediction. To assess training stability, the C-MAPSS trajectory visualizations and the corresponding internal stability summary were repeated with ten random seeds, and the reported statistics are given as mean $\pm$ standard deviation across runs. The controlled component analysis on XJTU reported later uses five runs to keep the module-level comparison computationally tractable. For the confidence-interval plots, the 95\% band at each time index is computed from the seed-wise empirical distribution as $(\mu_t \pm 1.96\sigma_t)$, capturing run-to-run training stochasticity rather than a full probabilistic uncertainty decomposition. The wider intervals in early life reflect higher trajectory ambiguity, whereas the narrower bands near end-of-life indicate stronger agreement among runs, which supports more reliable maintenance timing decisions \citep{LI202426}.

\begin{figure}[H]
  \centering
    \includegraphics[width=0.8\textwidth]{RUL_comparison_random_engine.png}
    \caption{RUL predictions for selected engines from the C-MAPSS dataset: Engine 38 (FD001), Engine 190 (FD002), Engine 50 (FD003), and Engine 131 (FD004). Solid blue lines denote the true RUL, orange dashed lines show predictions, and shaded regions represent 95\% confidence intervals.}

    \label{fig:rul_comparison}
\end{figure}

\subsubsection{XJTU dataset}
The dataset includes 55 lithium-ion batteries tested under six charging and discharging strategies. The proposed method is evaluated for SoH estimation, achieving superior results on several metrics \citep{WANG2024109884}.

As detailed in 
\autoref{tab:ComparisonOF-XJTU}, the proposed RGPD model demonstrates superior performance in estimating the SoH of lithium-ion batteries on the XJTU dataset, particularly excelling in MAPE, a critical metric for battery health assessment due to its relative error interpretability. With an average MAPE of 0.86\%, the proposed model outperforms Battery PINN (1.09\%), MLP (2.52\%), and CNN (2.88\%). It achieves lower MAPE in four out of six batches, with improvements ranging from 0.20\% to 0.48\%. While its average RMSE of 0.0105 is slightly higher than Battery PINN 0.0085, it remains highly competitive, especially in key batches, highlighting its robustness.

Although RGPD is a PINN-based model designed to handle diverse datasets, including bearings, engines, and batteries, it performs comparably to, and in some batches better than Battery PINN \citep{WANG2024109884}, which was specifically trained with physical constraints tailored solely to batteries. This comparison is particularly meaningful because battery degradation is governed mainly by electrochemical aging mechanisms, such as capacity fade and charge--discharge behavior, whereas turbofan engine degradation in C-MAPSS arises from mechanical and thermodynamic interactions among rotating components, operating conditions, and fault modes. Similarly, on the C-MAPSS dataset, RGPD also outperforms AttnPINN \citep{LIAO2023102195}, another PINN-based model developed for RUL prediction on the same benchmark, achieving a substantially lower average RMSE and Score. This demonstrates the strong generalizability of RGPD across different industrial prognostics applications. We emphasize, however, that this should not be interpreted as direct cross-domain transfer in the sense of training once on one asset family and deploying the same fitted model unchanged on another fundamentally different asset. Rather, the evidence supports methodological generality: the same design principles remain effective across intentionally heterogeneous PHM tasks. Taken together with the C-MAPSS and PHM2012 results, this suggests that the transferable part of the framework lies in its graph-based relational modeling and general degradation priors, while asset-specific adaptation mainly enters through preprocessing, graph construction, and failure-threshold definitions rather than through a different core architecture. This distinction is important for extending the framework to other artifact-centered engineering domains, where the sensing graph, target variable, and domain constraints must be defined according to the specific artifact and decision task.

\begin{table}[H]
    \centering
    \caption{Batch-wise MAPE and RMSE comparison of XJTU.}
    \label{tab:ComparisonOF-XJTU}
    \resizebox{\textwidth}{!}{%
    \begin{tabular}{l 
        *{6}{c c}
    }
        \toprule
        \multirow{2}{*}{Method}
            & \multicolumn{2}{c}{Batch\,1}
            & \multicolumn{2}{c}{Batch\,2}
            & \multicolumn{2}{c}{Batch\,3}
            & \multicolumn{2}{c}{Batch\,4}
            & \multicolumn{2}{c}{Batch\,5}
            & \multicolumn{2}{c}{Batch\,6} \\
        \cmidrule(lr){2-3} \cmidrule(lr){4-5} \cmidrule(lr){6-7}
        \cmidrule(lr){8-9} \cmidrule(lr){10-11} \cmidrule(lr){12-13}
        & RMSE     & MAPE     
        & RMSE     & MAPE     
        & RMSE     & MAPE     
        & RMSE     & MAPE     
        & RMSE     & MAPE     
        & RMSE     & MAPE\\     
        \midrule

        MLP     & 0.0260   & 0.0277   
                & 0.0275   & 0.0304   
                & 0.0211   & 0.0237   
                & 0.0200   & 0.0235   
                & 0.0183   & 0.0217   
                & 0.0204   & 0.0242   \\
        CNN     & 0.0270   & 0.0330   
                & 0.0298   & 0.0352   
                & 0.0177   & 0.0212   
                & 0.0150   & 0.0189   
                & 0.0350   & 0.0453   
                & 0.0149   & 0.0194   \\
        Battery PINN \citep{wang2024physics}    
                & \underline{0.0070}   & \underline{0.0094}   
                & \underline{0.0113}   & \underline{0.0122}   
                & \textbf{0.0086}   & \underline{0.0100}   
                & \textbf{0.0071}   & \textbf{0.0105}   
                & \textbf{0.0105}   & \textbf{0.0135}   
                & \textbf{0.0063}   & \underline{0.0097}   \\
\methodcell{\textbf{RGPD (Proposed)}}      
        & \bpmcell{0.0055}{0.0005}   & \bpmcell{0.0046}{0.0003}
        & \bpmcell{0.0100}{0.0004}   & \bpmcell{0.0086}{0.0007}
        & \upmcell{0.0093}{0.0006}   & \bpmcell{0.0080}{0.0004}
        & \upmcell{0.0134}{0.0010}   & \upmcell{0.0111}{0.0008}
        & \upmcell{0.0180}{0.0009}   & \upmcell{0.0137}{0.0012}
        & \upmcell{0.0070}{0.0005}   & \bpmcell{0.0055}{0.0006}  \\
        \bottomrule
    \end{tabular}%
    }
\end{table}

\begin{figure}[H]              
  \centering
  \includegraphics[width=\textwidth]{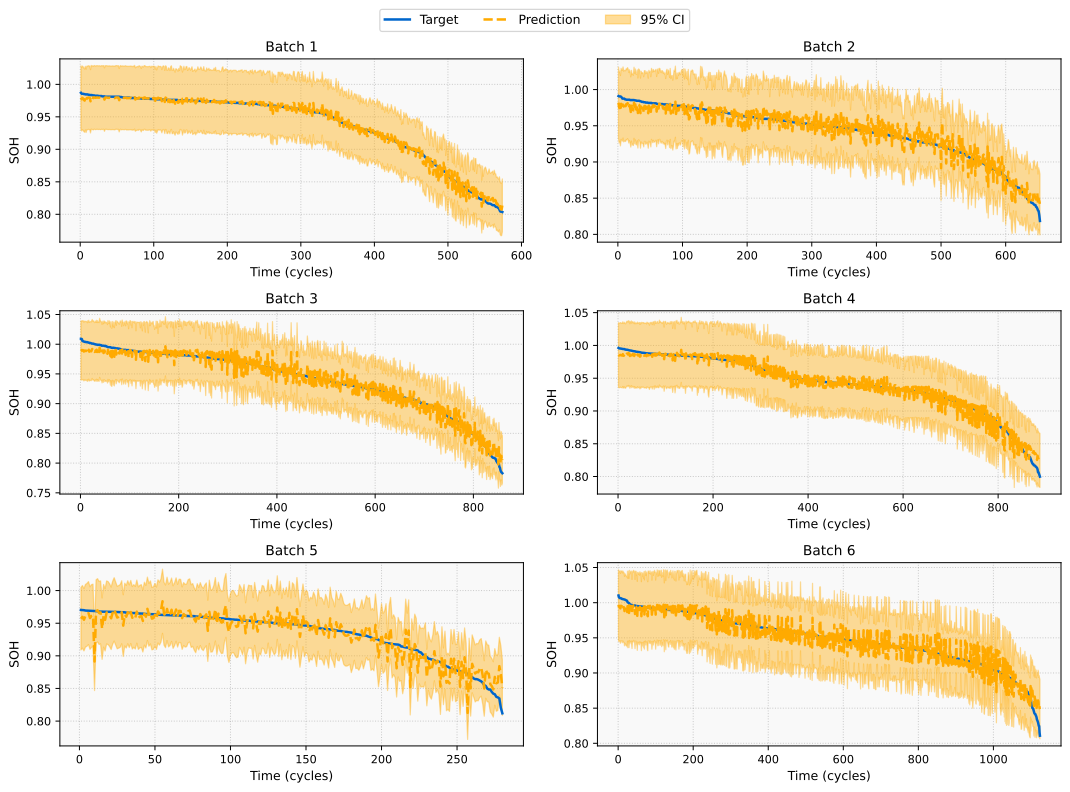}  
  \caption{SoH degradation trajectories for XJTU batches 1-6. Solid blue lines denote the true SoH targets, orange dashed lines show the predicted values, and shaded regions represent the 95\% confidence intervals. The model consistently follows the overall degradation trend, with wider uncertainty at early cycles that narrows toward end-of-life.}
  \label{fig:XJTU-SOH}

\end{figure}

As shown in \autoref{fig:XJTU-SOH}, the model successfully captures the overall SoH degradation trends across all six batches. Predictions generally follow the true targets, with wider confidence intervals in the early cycles indicating higher uncertainty, where the degradation signature is still weak and seed-wise predictions are more dispersed, and narrower intervals toward end-of-life reflecting improved reliability as the degradation pattern becomes more evident. In longer-life batches (e.g., Batches 4 and 6), predictions exhibit slightly larger fluctuations at the beginning but still align with the overall trend. By contrast, shorter-life batches (e.g., Batch 5) show tighter alignment with the true SoH, with stronger correlation and less deviation. These results highlight the model's ability to balance accuracy and uncertainty across varying degradation behaviors.

\subsubsection{PHM2012 Dataset}

\begin{figure}[H]                
  \centering
  \includegraphics[width=0.7\textwidth]{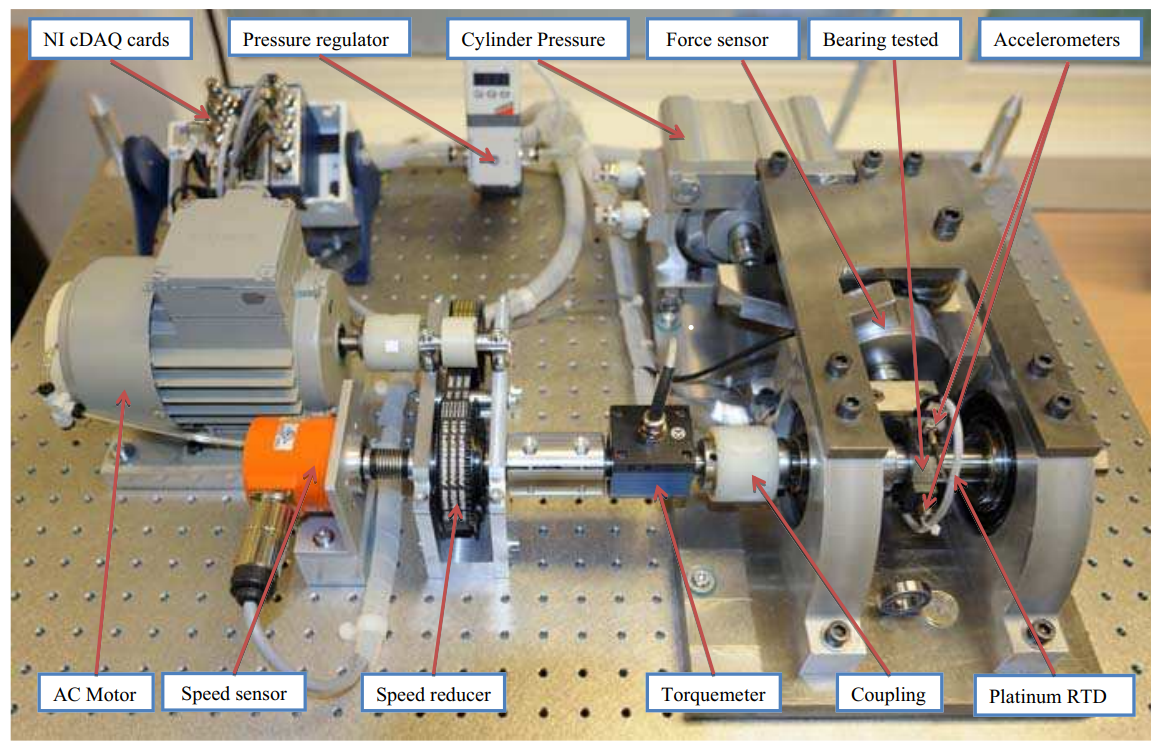}  
  \caption{Schematic of the PRONOSTIA experimental platform at FEMTO-ST, showing sensor placements for run-to-failure tests of deep-groove ball bearings of the PHM2012 dataset. Twin orthogonal accelerometers (25.6~kHz), a PT-100 temperature probe (10~Hz), and speed--torque--force channels (100~Hz) capture natural degradation under the operating conditions listed in \autoref{operating_conditions}.}
  \label{fig:PRONOSTIA}
\end{figure}

The PHM2012 bearing dataset \citep{nectoux2012pronostia}, introduced for the IEEE PHM 2012 Prognostics Challenge, provides run-to-failure measurements of deep-groove ball bearings under realistic operating conditions. The data were collected using the PRONOSTIA experimental platform at the FEMTO-ST laboratory in France, which is illustrated in \autoref{fig:PRONOSTIA}. This platform is instrumented with twin orthogonal accelerometers sampled at 25.6~kHz, a PT-100 temperature probe operating at 10~Hz, and speed, torque, and force sensors recorded at 100~Hz, enabling comprehensive monitoring of bearing degradation dynamics.

The dataset is organized into three distinct operating conditions, each defined by a specific combination of rotational speed and radial load. Across these conditions, a total of 17 complete run-to-failure experiments are provided for training, while 11 additional runs are truncated and reserved for testing and RUL evaluation. Notably, the first and second operating conditions each include seven bearing runs, whereas the third condition contains only three runs, reflecting a more constrained operating regime. The detailed operating parameters and the corresponding bearing identifiers for each condition are summarized in \autoref{operating_conditions}.

\begin{table}[H]
    \centering
    \small
    \setlength{\tabcolsep}{4pt} 
    \renewcommand{\arraystretch}{1.2}
    \caption{PHM2012 dataset: operating conditions and corresponding runs.
    The notation \textit{Bearing}$i$\_$j$ denotes the $j$-th bearing run under operating condition $i$.}
    \label{operating_conditions}
    \begin{tabular}{c r r l}
        \toprule
        Condition & Speed (rpm) & Load (N) & Dataset runs \\
        \midrule
        1 & 1800 & 4000 & Bearing1\_11\_7 \\
        2 & 1650 & 4200 & Bearing2\_12\_7 \\
        3 & 1500 & 5000 & Bearing3\_13\_3 \\
        \bottomrule
    \end{tabular}
\end{table}

Building upon this experimental setup, the PHM2012 challenge defines a fixed and standardized data partitioning strategy to ensure fair and reproducible evaluation of RUL prediction models. For each operating condition, a subset of bearing runs is assigned to the learning set, while the remaining runs are truncated and used exclusively as test data with hidden failure endpoints. This official split between training and testing data across all operating conditions is detailed in \autoref{operating_conditions2}.

\begin{table}[H]
    \centering
    \caption{PHM2012 data proportions.}
    \label{operating_conditions2}
    \renewcommand{\arraystretch}{1.1}
    \begin{tabular}{l c c c}
        \toprule
        Data sets & Condition 1 & Condition 2 & Condition 3 \\
        \hline
        \multirow{2}{*}{Learning set} 
        & Bearing1\_1 & Bearing2\_1 & Bearing3\_1 \\[0.1ex]
        & Bearing1\_2 & Bearing2\_2 & Bearing3\_2 \\[0.1ex]
        \hline
        \multirow{5}{*}{Test set} 
        & Bearing1\_3 & Bearing2\_3 & Bearing3\_3 \\[0.1ex]
        & Bearing1\_4 & Bearing2\_4 & -- \\[0.5ex]
        & Bearing1\_5 & Bearing2\_5 & -- \\[0.5ex]
        & Bearing1\_6 & Bearing2\_6 & -- \\[0.5ex]
        & Bearing1\_7 & Bearing2\_7 & -- \\[0.5ex]
        \hline
    \end{tabular}
\end{table}

Unlike datasets with artificially induced faults, the bearings in PHM2012 degrade naturally over time, often exhibiting concurrent defects in rolling elements, inner and outer races, and cages. This results in highly non-stationary and noisy signals that closely resemble real-world industrial degradation processes, making the dataset particularly challenging and representative for prognostics tasks.

In the preprocessing pipeline, raw horizontal acceleration signals are transformed into spectrogram-based representations to capture informative frequency-domain characteristics commonly used in time-series degradation analysis \citep{CORTESIBANEZ2020385}. Leveraging these representations, the proposed model is able to capture degradation trends across all bearings, including cases with sparse fault signatures or abrupt changes in degradation behavior.
To quantitatively assess the effectiveness of the proposed approach, its performance on the PHM2012 dataset is compared against several state-of-the-art methods using RMSE and MAE as evaluation metrics. The comparative results are reported in \autoref{PHM2012Comp}. As shown, the proposed RGPD model consistently achieves lower prediction errors than existing approaches, demonstrating superior accuracy and robustness across different operating conditions.
\begin{table}[H]
    \centering
    \caption{Performance comparison on PHM2012.}
    \label{PHM2012Comp}
    \begin{tabular}{l c c c}
        \toprule
        Model & Year & RMSE & MAE \\
        \midrule
        MCNN \citep{8384285} & 2019 & 0.1987 & 0.1542 \\
        COT \citep{9792382} & 2022 & 0.1096 & 0.0769 \\
        TT-ConvLSTM \citep{NIAZI2024110888} & 2023 & 0.1090 & 0.0870 \\
        WTE-Trans \citep{10105290} & 2024 & 0.1100 & 0.0930 \\
        ILCANet \citep{ZHANG2024110399} & 2025 & \underline{0.0890} & \underline{0.0670} \\
        \textbf{RGPD (Proposed)} & -- & \textbf{0.0778 $\pm$ 0.0032} & \textbf{0.0634 $\pm$ 0.0029} \\
        \bottomrule
    \end{tabular}
    \begin{flushleft}
    \footnotesize
    The best results are \textbf{bolded}, and the second-best results are \underline{underlined}. Literature baselines are reported as published values, while RGPD is reported as mean $\pm$ standard deviation over repeated runs.
    \end{flushleft}
\end{table}

\subsection{Missing Data}

For transparency, the simulated deletions were applied to the raw time-series prior to any feature-engineering or graph construction. To further assess robustness under harsher conditions and emulate real-world outages caused by hardware faults, communication errors, or maintenance issues, two deletion strategies were applied: (i) random missing values, where a fixed percentage of time-series entries was uniformly removed; and (ii) sensor-wise deletion, where entire feature channels were omitted per sample to simulate complete sensor failures. Three missingness levels were evaluated for both deletion strategies: $3.33\%$, $6.67\%$, and $10\%$. Prior to prediction, missing values were imputed using (1) linear interpolation along time for each feature, (2) mean imputation with feature-wise averages, or (3) zero imputation. To ensure reproducibility and fair comparison, missing-data masks were generated after train/validation/test splitting and applied independently within each sample window at the raw-signal level. During imputation, no future-cycle information was used: interpolation was performed only along available temporal points within each sample, and mean imputation used feature means estimated from the training split only. After imputation, the standard preprocessing and graph-construction pipeline was applied unchanged, so graph topology remained identical to the clean-data setting and only node attributes were affected by missingness. Across datasets, mean imputation provided the best accuracy-stability trade-off and is therefore the default strategy under moderate sensor loss in practical deployments.

\autoref{fig:SOAT-RMSE} and  \autoref{fig:SOAT-SCORE} demonstrate sustained predictive quality under both sparse losses and full-channel outages. Mean-based imputation consistently outperformed interpolation and zero-filling, with RMSE remaining below 20 and Score values in acceptable ranges. As expected, higher missingness produced moderate error increases, reflecting a clear, interpretable link between input corruption and output quality. Notably, for FD001 and FD003, mean imputation yielded performance comparable to the clean-data baseline and, in many cases, exceeded the average accuracy of competing methods, highlighting resilience in simpler settings. In FD002 and FD004, where multiple operating conditions and diverse fault modes introduce greater complexity, performance degradation was more evident as missingness increased. Still, under realistic scenarios such as sensor-wise deletion, the framework showed stronger robustness than in pointwise random deletion cases. While accuracy declined under high deletion ratios and added noise, results with mean imputation remained competitive with the average baseline performance reported in \autoref{Results_comparison_sorted_desc}. Overall, the framework demonstrates robustness to realistic sensor unavailability and reliable operation in imperfect industrial settings. This is particularly relevant for maintenance decision support, where sensor outages and incomplete monitoring records are common.

\begin{figure}[H]
    \centering
    \includegraphics[width=\textwidth]{missing_rmse_plots_paper_with_sota_rgpd_stuck.png}
    \caption{RMSE comparison between state-of-the-art models and RGPD across missingness levels.}
    \label{fig:SOAT-RMSE}
\end{figure}

\begin{figure}[H]
    \centering
    \includegraphics[width=\textwidth]{missing_score_plots_paper_with_sota_rgpd_stuck.png}
    \caption{Score comparison between state-of-the-art models and RGPD across missingness levels.}
    \label{fig:SOAT-SCORE}
\end{figure}

\subsection{Physical Constraints Across Case Studies}

\begin{figure}[H]
    \centering
    \includegraphics[width=1\linewidth]{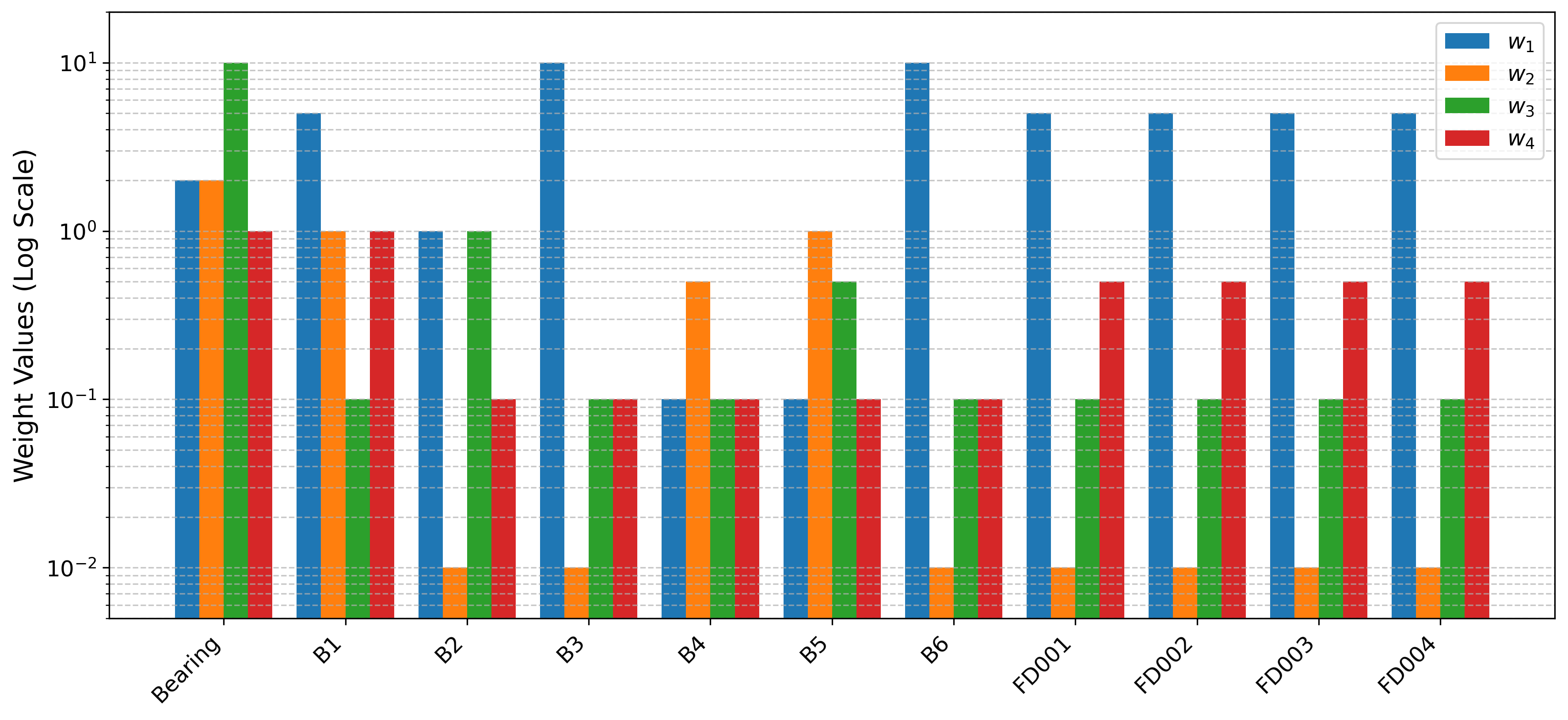}
    \caption{Learned weights across different prognostic datasets.}
    \label{Weights}
\end{figure}

As shown in \autoref{Weights}, Q-learning agents produced different weights for each physical constraint across lifetime data from diverse applications. In the C-MAPSS sub-datasets, the agents converged to identical weights, highlighting a strong connection between the relative importance of the constraints and their role in achieving accurate predictions, particularly in C-MAPSS. In XJTU, this connection is less pronounced, although a consistent trend between \(W_{3}\) (residual consistency) and \(W_{4}\) (broken-device penalty) is observed in most batches. Across all sub-datasets and applications, the decreasing RUL/SoH constraint \(W_{1}\) has consistently received greater emphasis than the others, underscoring its critical role in accurate RUL and SoH estimation.

\subsection{Visual Feature Contributions}

\begin{figure}[H]
    \centering
    \includegraphics[width=\linewidth]{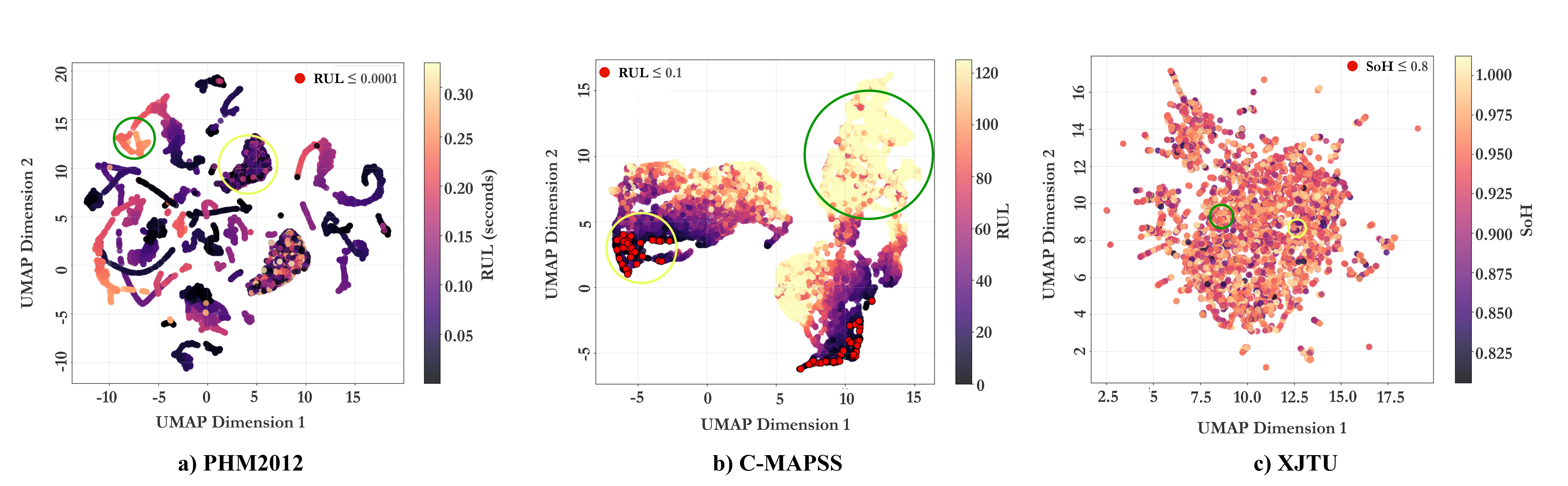}
    
    \caption{UMAP feature mapping for diverse lifetime datasets of (a) PHM2012, (b) C-MAPSS, and (c) XJTU. Green circle corresponds to healthy state and yellow circle corresponds to final degradation states.}
    \label{UMAPS}
\end{figure}

As shown in \autoref{UMAPS}, pannel (a) presents scattered degradation paths with distinct low-RUL clusters, highlighting
varied failure modes. This is reasonably caused due to presence of bearings with different operating conditions in the
evaluated dataset. In pannel (b), there is a clearer separation between healthy and degraded regions, and indicates
stronger RUL progression structures, because of the characterized and detailed features in long sequences of C-MAPSS.
pannel (c) ), presents a dense, less structured SOH space with subtle degradation patterns and weaker separability due to the critical difference of SOH and RUL, which the battery still has maintained at least 80\% of its initial capacity in
the end of its lifetime, so the features are mapped in a more overlapping state than RUL prediction.

\begin{figure}[H]
    \centering
    \includegraphics[width=1\linewidth]{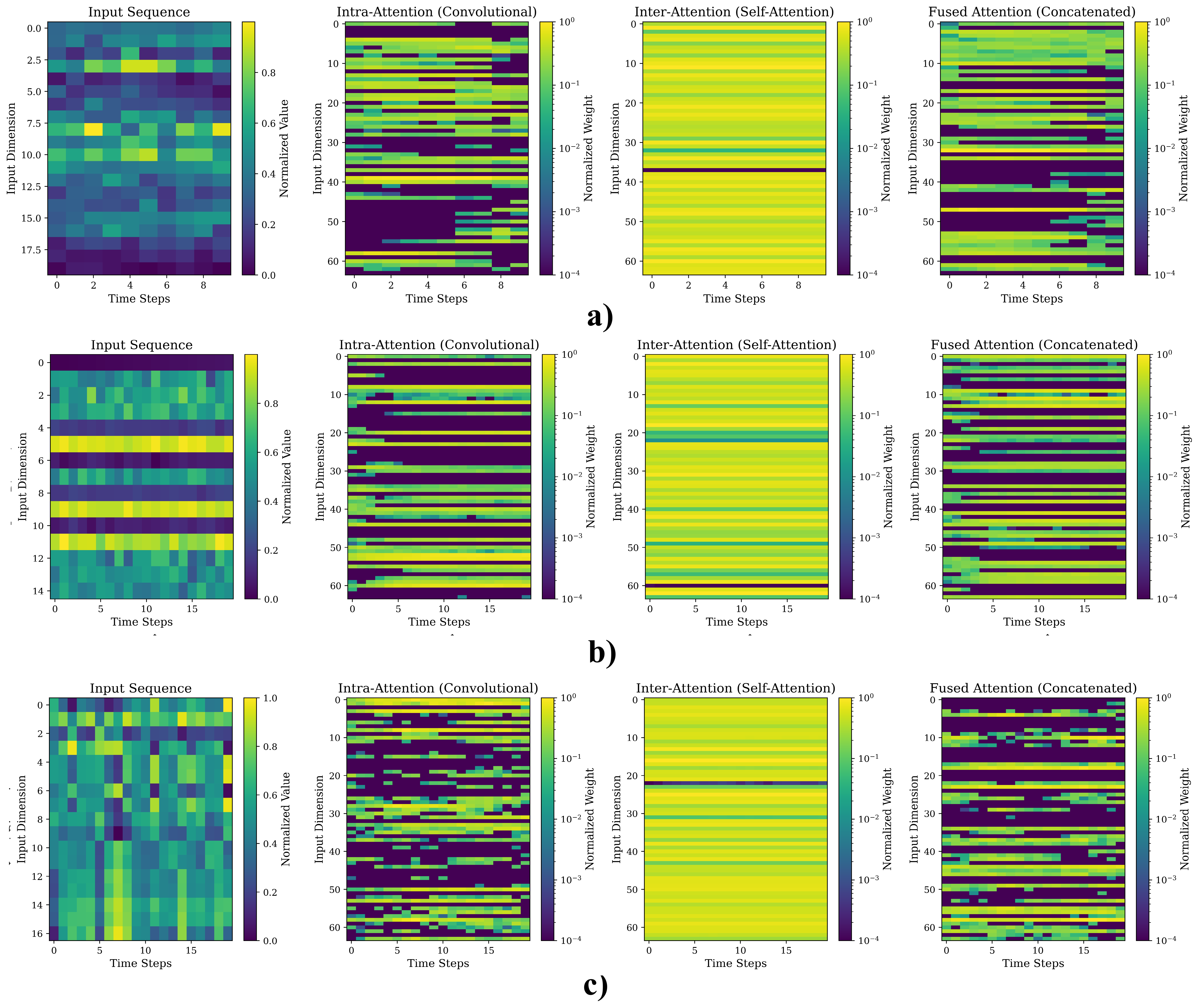}
    \caption{Attention visualization for (a) PHM2012, (b) XJTU, and (c) C-MAPSS. Columns show input sequences, intra-attention (convolutional), inter-attention (self-attention), and fused attention. Intra-attention emphasizes local dependencies, inter-attention captures global relationships, and fused attention integrates both for richer representations.}
    \label{fig:attentionweight}
\end{figure}

As explained in the methodology, TAU consists of two components: inter-attention (self-attention) and intra-attention (convolutional). In  \autoref{fig:attentionweight}, the effect of these two attention mechanisms and their fusion on the input sequences is illustrated for samples from PHM2012, XJTU, and C-MAPSS. Each row shows the raw input sequence, intra-attention, inter-attention, and fused attention. Intra-attention highlights localized and short-term dependencies with sparse, high-contrast activations, while inter-attention provides a more uniform distribution, capturing global and long-range relationships. Fused attention combines both perspectives, yielding richer and more balanced feature representations that integrate local details with global context.

\subsection{Ablation Study}
For a better understanding of each component effects, the model is tested with and without each component across all datasets. 

\begin{table}[H]
  \centering
  \caption{Performance metrics of ablation models on PHM2012. (Training time is reported per epoch; inference time is reported per test bearing .)}
  \label{tab:bearing-performances}
  \renewcommand{\arraystretch}{1.6}
  \begin{tabular}{lccccc}
    \toprule
    Model & RMSE & MAE & Train time (s) & Test time (s) & Parameters \\
    \midrule
    Complete model      & \textbf{0.0778} & \textbf{0.0634} & 2.52 & 0.095 & 250,681 \\
    
    Without RL          & 0.0923 & 0.0775 & \textbf{2.01} & \textbf{0.082} & \textbf{229,557} \\
    
    
    Without TAU         & 0.0818 & 0.0681 & 2.04 & 0.092 & 239,593\\
    \bottomrule
  \end{tabular}
\end{table}

\begin{table}[H]
  \centering
  \caption{Performance metrics of ablation models on C-MAPSS subsets. (Training time is reported per epoch; inference time is reported per test engine.)}
  \label{tab:C-MAPSS-Performances}
  \renewcommand{\arraystretch}{1.5}
  \resizebox{\textwidth}{!}{
  \begin{tabular}{l cc cc cc cc ccc}
    \toprule
    \multirow{2}{*}{Model} 
      & \multicolumn{2}{c}{FD001}
      & \multicolumn{2}{c}{FD002}
      & \multicolumn{2}{c}{FD003}
      & \multicolumn{2}{c}{FD004} 
      & \multicolumn{3}{c}{Complexity}\\
    \cmidrule(lr){2-3} \cmidrule(lr){4-5}
    \cmidrule(lr){6-7} \cmidrule(lr){8-9}
    \cmidrule(lr){10-12}
      & RMSE & Score 
      & RMSE & Score 
      & RMSE & Score 
      & RMSE & Score
      & Train time (s) & Test time (ms) & Parameters\\
    \midrule
    Complete model      & \textbf{10.771} & \textbf{207.09}  & 11.58 &  684.24 & \textbf{10.785} & \textbf{245.28}  & 12.381 & 787.39  & 10.42 & 0.063 & 250,731\\
    
    Without RL          & 11.556 & 290.15  & 11.477 & 653.07  & 10.804 & 274.41  & 12.630 & 781.30 & \textbf{9.97} & \textbf{0.061} & \textbf{229,557}\\
    
    
    Without TAU         & 12.334 & 320.10  & 12.209 & 851.67  & 11.092 & 311.25  & 13.031 & 904.63 & 11.06 & 0.074 & 272,281\\
    \bottomrule
  \end{tabular}
  }
\end{table}

\begin{table}[H]
  \centering
  \caption{Performance metrics of ablation models on different batches of XJTU. (Training time is reported per epoch; inference time is reported as the average across test batches.)}
  \label{tab:battery-performances}
  \renewcommand{\arraystretch}{1.4}
  \resizebox{\textwidth}{!}{
  \begin{tabular}{l c c c c c c c c c c c c c c}
    \toprule
    \multirow{2}{*}{Model} 
      & Batch1 & Batch2 & Batch3 & Batch4 & Batch5 & Batch6 
      & \multicolumn{3}{c}{Complexity}\\
    \cmidrule(lr){8-10}
      & RMSE & RMSE & RMSE & RMSE & RMSE & RMSE 
      & Train time (s) & Test time (s) & Parameters \\
    \midrule
    Complete model      
      & \textbf{0.0055} & 0.0100 & \textbf{0.0093} & 0.0134 & 0.0180 & \textbf{0.0070} 
      & 2.54 & 0.110 & 359,353\\
      
    Without RL          
      & 0.0057 & \textbf{0.0069} & 0.0096 & \textbf{0.0103} & \textbf{0.0117} & 0.0112 
      & \textbf{2.27} & \textbf{0.098} & \textbf{338,229}\\


    Without TAU         
      & 0.0568 & 0.0212 & 0.2896 & 0.1365 & 0.0884 & 0.0084 
      & 2.90 & 0.123 & 381,481\\
    \bottomrule
  \end{tabular}
  }
\end{table}

This section presents ablation of some components, creating 
two cases: removing RL agents
, and the TAU block. Each ablation helps isolate the contribution of individual modules to the overall model performance. In  
\autoref{tab:bearing-performances}, \autoref{tab:C-MAPSS-Performances},
and
\autoref{tab:battery-performances},
the ablation results from PHM2012, C-MAPSS, and XJTU are listed, respectively.

\subsubsection{Complexity Analysis}
The reported timing values are not exact due to randomness and other factors. Execution times can also vary even when using the same GPU. Nevertheless, the results show that the proposed model requires slightly more computation time compared to the baseline models, while the inference times remain almost identical. This makes the additional training cost worthwhile, as it is a one-time process to obtain a trained model.

Moreover, the number of parameters introduced by the RL component is small, and the overall size of the main model is not excessively large. Therefore, it can be executed even on low-end GPUs. The reported execution times are for processing the entire dataset; the time per single sample is significantly lower.
It is also important to note that parallel processing and advanced acceleration settings were not utilized. For instance, the \texttt{num-workers} parameter was not configured, which could otherwise significantly accelerate data loading.
Although the model combines several components such as graph networks and reinforcement learning, which may suggest high computational complexity, relatively small network architectures were employed in each component. This design choice prevents the model from becoming prohibitively large or computationally expensive.

It is worth noting that, in testing without TAU, a slightly larger attention module was used in most cases. However, results in terms of RMSE, score, computation time, and model parameters indicate that TAU is more effective for this task, even with fewer parameters. 
In addition, the reinforcement learning component introduces only a small number of extra parameters compared to the base model, leading to a marginal overhead during training and a negligible impact on inference time. This trade-off is considerably more efficient than dynamically searching for hyperparameters.

\subsubsection{Reinforcement Learning Role}
The model can adaptively assign weights to the physics-informed loss terms and dynamically scale hidden representations, thanks to the RL modules, especially the Q-learning agents. This dynamic adjustment is essential for striking a balance between physical consistency and empirical data fitting, and leads to higher accuracy and generalization, particularly when working with diverse datasets that have different degradation dynamics.

The RL modules play two complementary roles in the proposed framework: Q-learning rebalances the physics-informed losses online, while SAC modulates latent features before TAU. Across C-MAPSS and PHM2012, removing RL consistently degrades accuracy, indicating that adaptive weighting and feature modulation are particularly useful when the degradation process is heterogeneous or noisy. On XJTU, the effect is more nuanced. The targeted XJTU study reported later in this section shows that removing SAC can slightly improve validation RMSE, whereas removing dynamic Q-weighting increases the mean test RMSE. This suggests that RL is most valuable not as a guaranteed way to lower the training loss in every case, but as a mechanism for improving generalization and balancing competing objectives under cross-condition variability. In smaller or smoother settings, the gains can be narrower, which is consistent with the additional configuration cost introduced by RL and motivates the controlled tuning comparison reported below.

In the proposed method, SAC learns a scaling vector applied to the GCRN hidden features to stabilize the input of TAU. The reward function is defined as -MSE; thus, any error in the final prediction reduces the reward. If informative features are overly suppressed or noise is amplified, the error increases, and SAC adjusts the scales accordingly. This naturally balances noise reduction and feature preservation. To avoid bias, the scaling is restricted within predefined bounds and adds a penalty that keeps scales close to 1. Potential bias may occur under noisy labels or distribution shifts, which can be mitigated through data augmentation and physics-informed constraints.

\subsubsection{Temporal Attention Unit Role}
In this section, the TAU blocks of the model are replaced with simple multi-head attention layers. As evident from the tables, among the components analyzed, TAU proves to be the most impactful for maintaining predictive accuracy. It enables the model to focus on spatio-temporal features, and dynamically adjusts the weight of past observations in recurrent processing. This capability is crucial for modeling global and local dependencies in degradation sequences.
Unlike standard self-attention, which treats temporal inputs as a uniform sequence, TAU explicitly disentangles intra-time correlations (relations among sensors or variables at the same timestamp) from inter-time dynamics (temporal evolution across timesteps). 
This separation allows the model to better capture both instantaneous sensor interactions and long-term degradation trends, which are central to industrial time-series data such as batteries and bearings.

The removal of TAU results in the steepest decline in performance across all datasets. In C-MAPSS, models without TAU perform worst on all four subsets, with FD004 suffering the most, likely due to its longer sequences and higher variance in operating conditions. 
In the XJTU dataset, the ablation of TAU leads to dramatic increases in RMSE and MAE, particularly in Batches 4 and 6. These results affirm that TAU is essential for tracking gradual and non-linear degradation, especially where degradation signals are weak or noisy. Without it, the model struggles to discriminate between informative and redundant signals, which leads to predictions with lower accuracy.

\subsubsection{Targeted Component and Tuning Analysis}

To obtain clearer module-level attribution on the XJTU lithium-ion battery dataset, we conducted two controlled analyses summarized in Table~\ref{tab:extended_ablation_xjtu}. Panel~(a) evaluates the contribution of the main architectural and physics-guided components. The largest degradation occurs when the graph-based backbone is replaced by BiLSTM, increasing the test RMSE from 0.00935 to 0.01333. This shows that the gain is not simply due to temporal sequence modeling, but mainly comes from explicitly capturing sensor interactions through the graph-based spatio-temporal backbone. For the XJTU battery data, where multiple measured variables evolve jointly during degradation, preserving this relational structure is more effective than treating the input as a purely sequential signal.

The physics-informed terms also improve performance, although their effect is more moderate than that of the graph backbone. Removing them increases the test RMSE from 0.00935 to 0.00966, suggesting that they primarily act as regularizers that improve physical consistency and robustness. The No SAC Scaling variant gives the best validation RMSE but a worse test RMSE, indicating a validation--test mismatch. This suggests that SAC scaling mainly supports generalization rather than in-sample fitting. Overall, Panel~(a) indicates that the full RGPD model benefits from the combination of graph-based representation learning, physics-guided regularization, and scaling-based stabilization.

Panel~(b) examines how the physics-informed loss-balancing mechanism should be controlled. Since the proposed model uses four bounded Q-learning weights, the comparison focuses on alternative strategies for selecting or adapting the same set of physics-related weights under a comparable training setting. The fixed-weight baseline remains competitive and has the lowest per-epoch cost, showing that static weights can provide a practical baseline. However, its higher test error indicates that fixed weights cannot fully capture the changing importance of different loss terms during training.

The MLP-based dynamic weighting strategy performs worse than the proposed Q-weighting approach despite its flexibility, suggesting that directly learning continuous loss weights may introduce instability or overfitting in this prognostics setting. Random search is also sub-optimal, indicating that the weight space is sensitive to uninformed static sampling. Bayesian TPE improves over random search but remains inferior to RGPD while requiring much higher computational cost. Thus, the advantage of Q-weighting is not only reduced manual tuning, but also efficient and stable adaptation of multiple physics-guided objectives.

Taken together, these experiments show that RGPD achieves its best performance through complementary mechanisms. The graph backbone captures sensor-level dependencies, the physics-informed losses constrain predictions toward plausible degradation behavior, SAC scaling improves generalization, and Q-weighting adaptively balances competing physical objectives. This combination yields the lowest test error with modest computational overhead compared with search-based tuning strategies.

\begin{table}[H]
\centering
\caption{Targeted component and weight-tuning analysis on the XJTU dataset. Panel~(a) reports mean $\pm$ standard deviation over five runs for controlled component variants. Panel~(b) compares alternative physics-weight tuning strategies. Training time is reported per epoch; the results are reported based on the average between 6 XJTU batches.}
\label{tab:extended_ablation_xjtu}
\renewcommand{\arraystretch}{1.2}
\small
\begin{tabular}{lccc}
\toprule
\multicolumn{4}{l}{\textbf{(a) Controlled component analysis}} \\
\midrule
\textbf{Model Variant} & \textbf{Valid RMSE} & \textbf{Test RMSE} & \textbf{Train / epoch (s)} \\
\midrule
Full RGPD (Proposed) & 0.00912 $\pm$ 0.00072 & \textbf{0.00935 $\pm$ 0.00078} & 2.54 \\
No SAC Scaling & \textbf{0.00874 $\pm$ 0.00012} & 0.01021 $\pm$ 0.00147 & 2.37 \\
No Physics (Supervised Only) & 0.00944 $\pm$ 0.00061 & 0.00966 $\pm$ 0.00069 & 1.98 \\
BiLSTM Backbone & 0.01282 $\pm$ 0.00040 & 0.01333 $\pm$ 0.00057 & \textbf{1.70} \\
\addlinespace[4pt]
\multicolumn{4}{l}{\textbf{(b) Weight-tuning strategies}} \\
\midrule
\textbf{Strategy} & \textbf{Valid RMSE} & \textbf{Test RMSE} & \textbf{Train / epoch (s)} \\
\midrule
Full RGPD (Proposed) & \textbf{0.00912 $\pm$ 0.00072} & \textbf{0.00935 $\pm$ 0.00078} & 2.54 \\
No Dynamic Q (Fixed Weights) & 0.00991 $\pm$ 0.00048 & 0.01018 $\pm$ 0.00062 & \textbf{2.34}  \\
MLP dynamic weights & 0.01101 $\pm$ 0.00075 & 0.01155 $\pm$ 0.00092 & 2.67 \\
Bayesian TPE static best & 0.01108 $\pm$ 0.00068 & 0.01155 $\pm$ 0.00085 & 27.10 \\
Random search static best & 0.01166 $\pm$ 0.00095 & 0.01225 $\pm$ 0.00110 & 2.45 \\
\bottomrule
\end{tabular}
\normalsize
\end{table}

\subsubsection{Hyperparameter Sensitivity}

\begin{figure}[H]
\centering
\includegraphics[width=\textwidth]{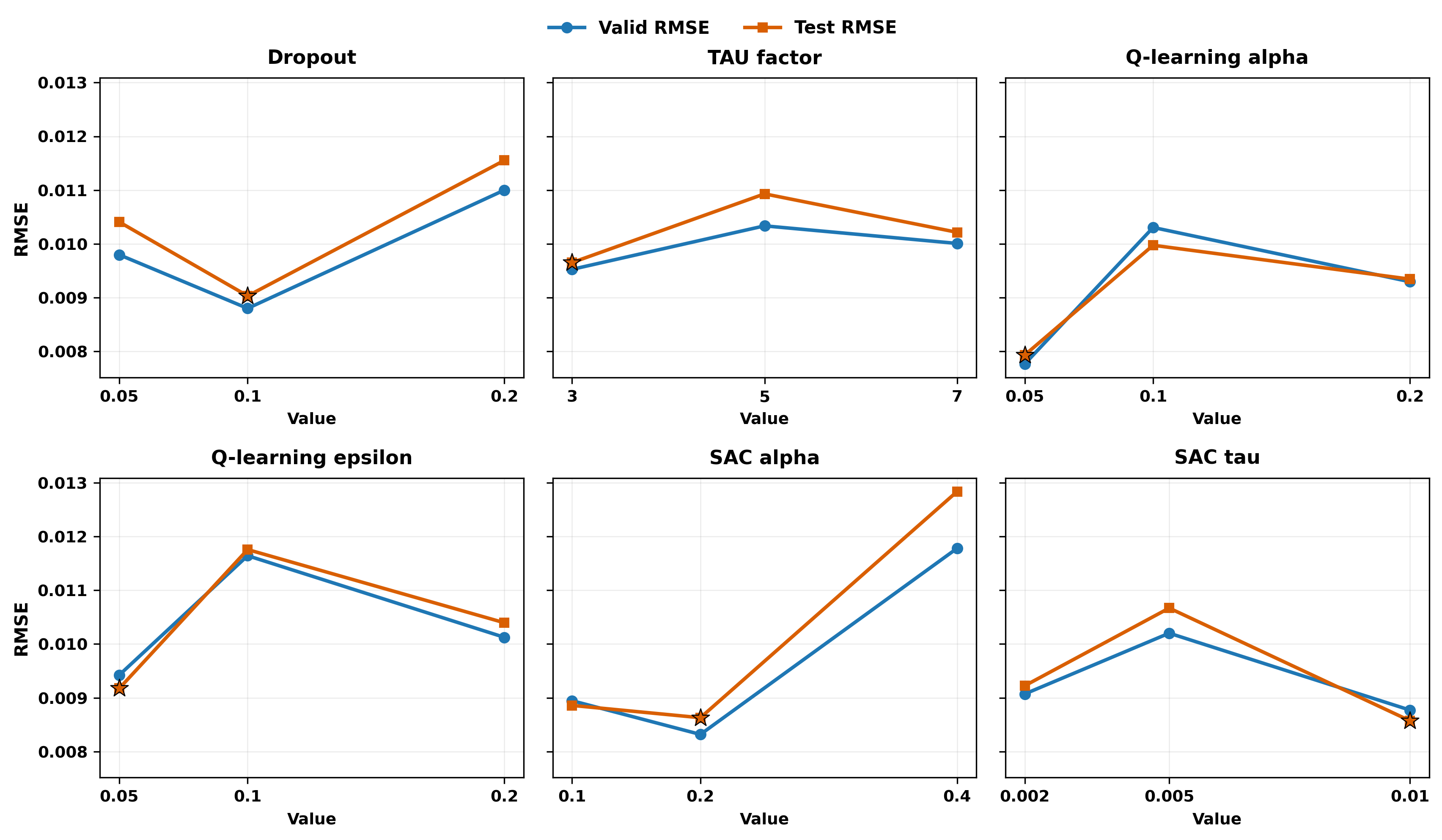}
\caption{One-factor-at-a-time hyperparameter sensitivity analysis on the XJTU dataset. Each panel varies one hyperparameter while keeping the remaining settings fixed and reports validation and test RMSE. Lower RMSE indicates better performance.}
\label{fig:xjtu_hparam_sensitivity}
\end{figure}

To assess the stability of RGPD with respect to key configuration choices, we conducted a one-factor-at-a-time (OFAT) sensitivity analysis on the XJTU dataset. In this analysis, one hyperparameter was varied at a time while all remaining settings were kept fixed. The evaluated parameters include dropout, which controls neural regularization; TAU factor, which scales temporal features; Q-learning $\alpha$ and $\epsilon$, which determine Q-table update strength and exploration in physics-weight selection; and SAC $\alpha$ and $\tau$, which control entropy regularization and target-network soft updates in the SAC controller. Figure~\ref{fig:xjtu_hparam_sensitivity} reports the corresponding validation and test RMSE values.

The RGPD exhibits differential robustness across hyperparameters. Moderate dropout (0.1) yields the most favorable performance, with degradation at both lower and higher levels, indicating an optimal regularization sweet spot. In contrast, the model demonstrates notably high sensitivity to SAC $\alpha$, where intermediate values produce a stable low-error region, but performance deteriorates sharply at higher entropy regularization (0.4), particularly on the test set. Stronger effects are also observed for the TAU factor, Q-learning $\alpha$ and $\epsilon$, and SAC $\tau$, which display pronounced non-monotonic behaviors (including inverted U-shapes), underscoring the need for careful tuning of temporal scaling and reinforcement learning adaptation parameters. Overall, while the architecture shows reasonable robustness to several configuration choices, its performance is most vulnerable to parameters governing exploration-exploitation balance and target network dynamics. As a single-seed OFAT screening, these results highlight relative sensitivity trends to inform subsequent multi-run hyperparameter optimization rather than providing definitive optima.

\section{Conclusion}

In this study, we proposed RGPD, a unified prognostics framework for RUL and SoH prediction across heterogeneous PHM systems. The framework combines graph-based spatio-temporal representation learning with physics-informed degradation constraints whose contributions are adaptively controlled during training. This design allows the model to capture multi-sensor degradation dependencies while maintaining physically plausible prediction trajectories.

Experimental results on C-MAPSS, PHM2012, and XJTU demonstrate the effectiveness of the proposed framework across turbofan engines, bearings, and lithium-ion batteries. The model improves the average RMSE by 12\% on the benchmark evaluations and achieves a 20\% lower MAPE on the XJTU dataset compared with the second-best baseline. These results show that the proposed design improves prediction accuracy across different degradation settings rather than only on a single asset type.

Beyond the reported accuracy gains, the key methodological contribution is the coordinated use of three mechanisms: graph-based spatio-temporal modeling for sensor dependency learning, SAC-based feature modulation for suppressing noisy latent representations, and Q-learning-based weighting for balancing physics-informed constraints. In particular, Q-learning turns the PINN loss weights into adaptive training decisions, allowing monotonicity, smoothness, failure-state consistency, and latent-dynamics constraints to be balanced without exhaustive manual tuning. This combination provides a reusable framework for heterogeneous PHM problems while preserving the flexibility needed for dataset-specific degradation behavior.

At the same time, this flexibility comes with a practical trade-off. Compared with simpler prognostic models, RGPD introduces additional training and implementation overhead through the joint use of graph-based temporal learning, reinforcement-driven feature modulation, and physics-informed loss balancing. This added complexity is most justified when multi-sensor dependencies, noisy latent representations, and multiple degradation constraints must be handled simultaneously. In more resource-constrained deployments or smaller datasets, lighter configurations, such as fixed physics weights or selectively simplified adaptive modules, may offer a practical balance between accuracy, interpretability, and computational efficiency.

Future work will focus on strengthening the physics-informed component by extending the current asset-agnostic residuals with asset-aware physics terms defined from domain-relevant state variables, operating descriptors, and failure indicators. Such extensions would preserve the graph-temporal learning backbone while tailoring the physics-informed residuals to domain-specific degradation mechanisms. Another important direction is validation in real-world online industrial settings, where continual calibration of latent degradation dynamics can be evaluated under streaming data, changing operating conditions, and deployment constraints.

More broadly, RGPD contributes to knowledge-intensive engineering decision support by combining sensor-level evidence, degradation priors, and adaptive loss control into a unified prognostic representation. Rather than replacing human or system-level maintenance policies, the framework provides more reliable and responsive health-state estimates that can inform RUL/SoH-based maintenance planning.

\bibliographystyle{unsrtnat}
\bibliography{references}

\appendix
\section{Details of the Graph Attention Convolutional Recurrent Network}
\label{appendix:GATGCRN}

\subsection{Graph Attention Convolution}

\begin{figure}[H]
  \centering
  \includegraphics[width=0.55\textwidth]{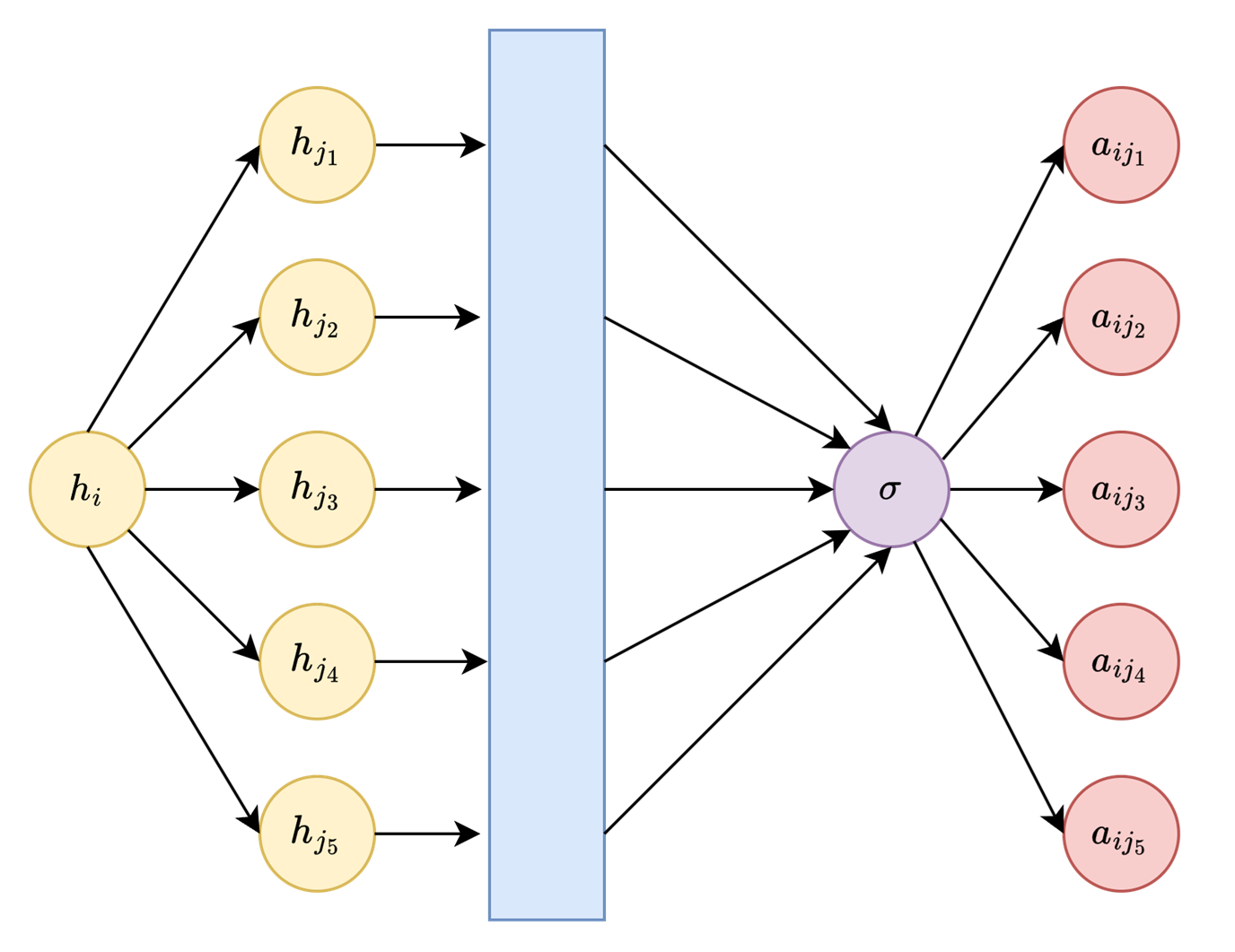}
  \caption{Structure of the GATConv layer. The attention mechanism adaptively weights neighboring nodes, allowing the model to focus on the most informative relationships within the graph.}
  \label{fig:GATConv}
\end{figure}

The GATConv layer extends traditional graph convolutions by introducing an attention mechanism that learns the importance of each neighbor dynamically.  
For a node \( i \) and its neighbor \( j \), the unnormalized attention coefficient is computed as:
\[
e_{ij} = \text{LeakyReLU}(a^\top [W h_i \| W h_j]),
\]
where \( W \) is a learnable weight matrix and \( a \) is a shared attention vector.  
Normalized attention coefficients are obtained using a softmax function:
\[
\alpha_{ij} = \frac{\exp(e_{ij})}{\sum_{k \in \mathcal{N}_i} \exp(e_{ik})},
\]
and node features are updated through attention-weighted aggregation:
\[
h_i' = \sigma\left(\sum_{j \in \mathcal{N}_i} \alpha_{ij} W h_j \right).
\]
For multi-head attention with \( K \) heads, outputs are averaged:
\[
h_i^{(l+1)} = \sigma\left(\frac{1}{K}\sum_{k=1}^{K}\sum_{j \in \mathcal{N}_i}\alpha_{ij}^k W^k h_j^{(l)}\right).
\]
This mechanism enables the model to emphasize significant spatial relationships and reduce the influence of weak or noisy connections.

\subsection{Graph Convolutional Recurrent Network}

\begin{figure}[H]
  \centering
  \includegraphics[width=0.7\textwidth]{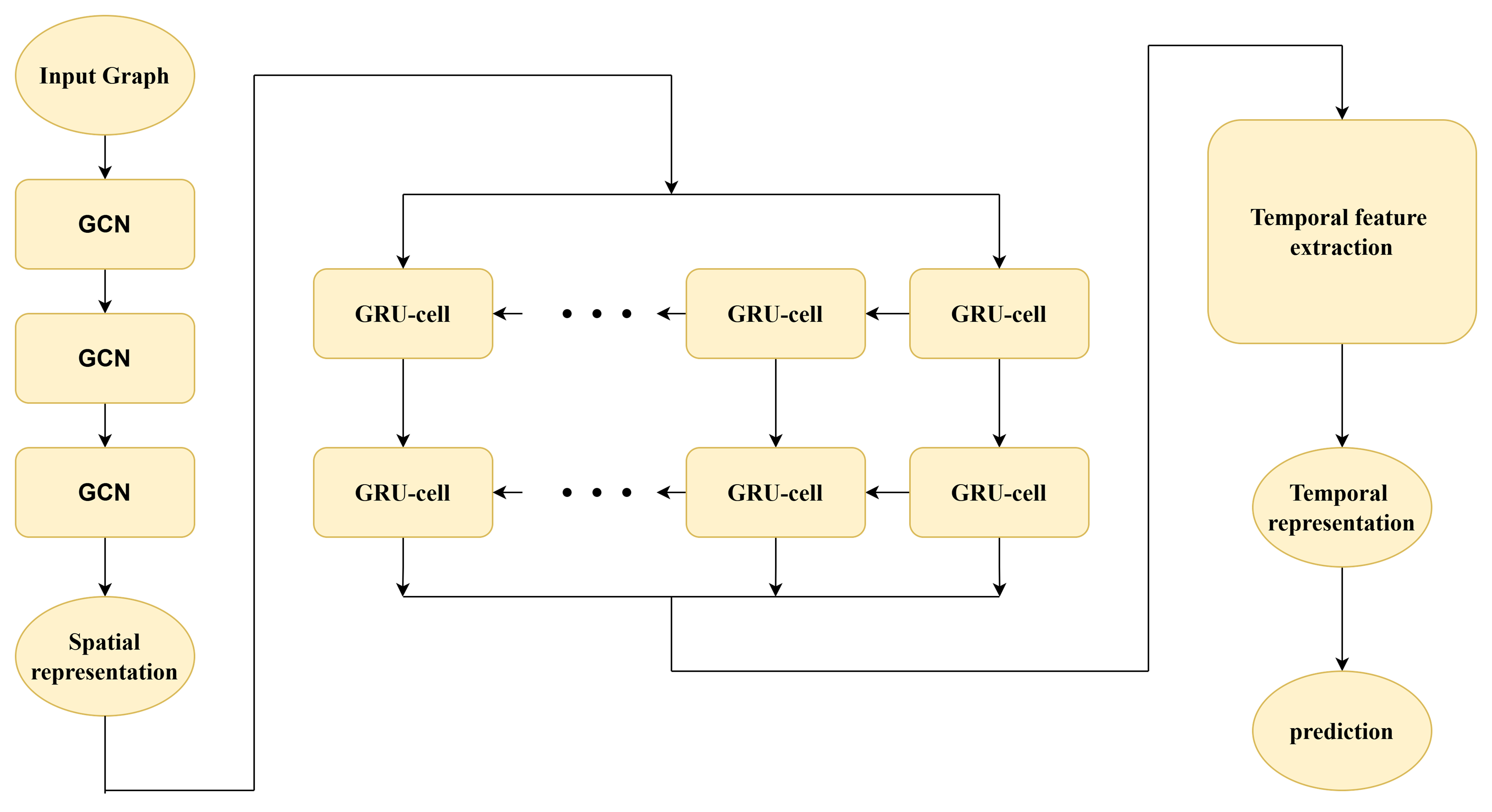}
  \caption{GCRN block. The architecture combines graph convolutions for spatial feature extraction with recurrent processing for temporal modeling.}
  \label{fig:GCRNN}
\end{figure}

The GCRN integrates graph convolutions into a recurrent structure to capture spatial and temporal dependencies jointly.  
At each time step \( t \), the hidden state update is expressed as:
\[
H_t = f_{\text{RNN}}(X_t, \text{GConv}(H_{t-1})),
\]
where \( X_t \) is the input at time \( t \), \( H_t \) is the hidden state, and \( \text{GConv}(\cdot) \) denotes the graph convolution operator.  
In this work, the recurrent unit is implemented using a GRU, which efficiently captures temporal dependencies while maintaining computational simplicity.

\subsection{Gated Recurrent Unit}

\begin{figure}[H]
  \centering
  \includegraphics[width=0.7\textwidth]{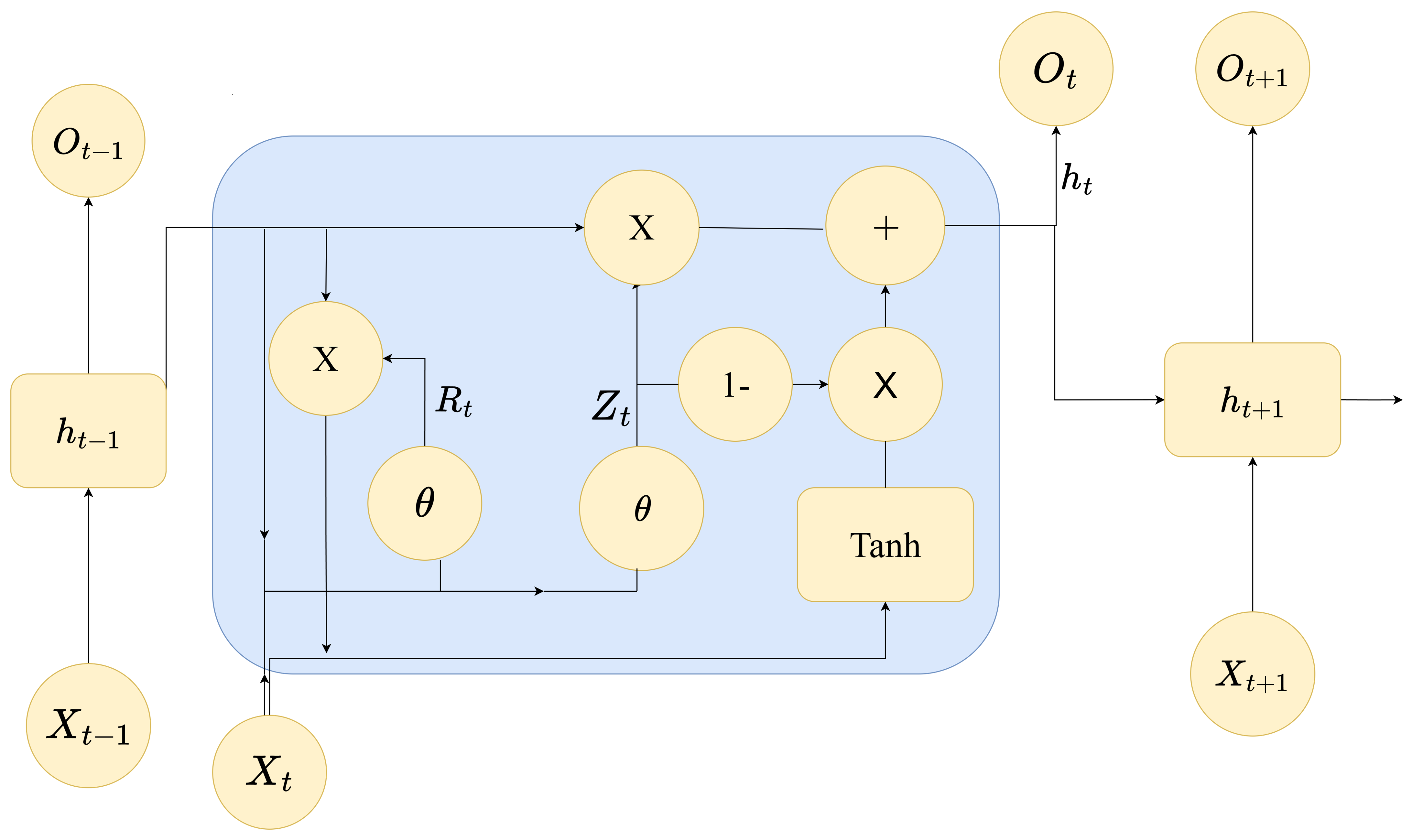}
  \caption{The update and reset gates regulate the flow of temporal information, enabling efficient modeling of long-term dependencies.}
  \label{fig:GRU}
\end{figure}

The GRU updates its hidden state through gated mechanisms as follows:
\[
z_t = \sigma(W_z X_t + U_z H_{t-1}), \quad
r_t = \sigma(W_r X_t + U_r H_{t-1}),
\]
\[
\tilde{H}_t = \tanh(W_h X_t + r_t \odot (U_h H_{t-1})), \quad
H_t = (1 - z_t) \odot H_{t-1} + z_t \odot \tilde{H}_t,
\]
where \( z_t \) and \( r_t \) denote the update and reset gates, respectively, and \( \odot \) represents element-wise multiplication.  
The GRU facilitates temporal learning by adaptively controlling the influence of past hidden states, thereby maintaining stable long-term dependencies.

\subsection{Model Configuration and Design Choice}

In the proposed GAT-GCRN model, the GATConv layer precedes the GCRN to ensure that spatial features are refined before temporal modeling. The attention mechanism in GATConv allows the model to emphasize key structural connections, while the GCRN captures the temporal evolution of these relationships across sequential data.  
All graph convolutions employ an adjacency matrix with self-loops to preserve node identity information. Multi-head attention is adopted to stabilize training and improve representation robustness, and GRU cells are used within the GCRN to balance accuracy and computational efficiency.

This architecture allows the model to simultaneously learn spatial and temporal patterns in complex graph-based time series, improving predictive performance and interpretability.

\section{Details of the Soft Actor-Critic}
\label{appendix:SAC}

\begin{figure}[H]
  \centering
  \includegraphics[width=0.75\textwidth]{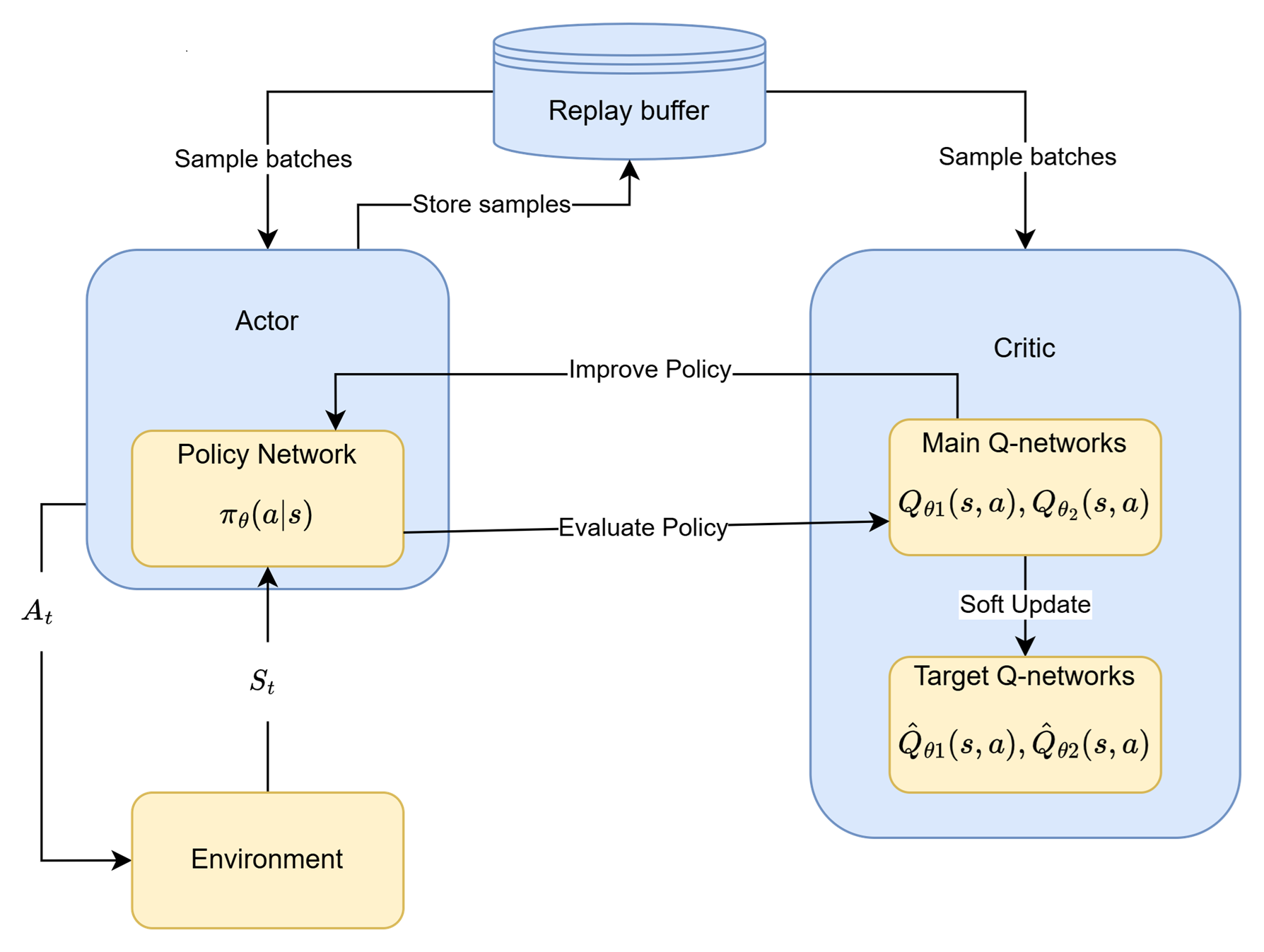}
  \caption{Schematic representation of the SAC framework.}
  \label{fig:SAC_appendix}
\end{figure}

SAC optimizes a stochastic policy by jointly maximizing the expected cumulative reward and policy entropy:
\[
J(\pi) = \mathbb{E}_{\pi}\!\left[\sum_{t=0}^{T} \gamma^t \big(r(s_t, a_t) + \alpha \mathcal{H}(\pi(\cdot | s_t))\big)\right],
\]
where \( \gamma \) is the discount factor, \( \alpha \) controls the entropy term, and 
\( \mathcal{H}(\pi(\cdot | s_t)) = -\mathbb{E}_{a_t \sim \pi}[\log \pi(a_t|s_t)] \) promotes exploration.

The critic network estimates the soft Q-value via the modified Bellman operator:
\[
\mathcal{T}^{\pi} Q(s_t, a_t) = r(s_t, a_t) + \gamma \mathbb{E}_{s_{t+1} \sim p}\!\left[V(s_{t+1})\right],
\]
with the value function defined as:
\[
V(s_t) = \mathbb{E}_{a_t \sim \pi}\!\left[ Q(s_t, a_t) - \alpha \log \pi(a_t | s_t) \right].
\]

The optimization losses are expressed as:
\[
L_{\text{critic}} = \text{MSE}\!\left(Q(s_t, a_t),\, r(s_t, a_t) + \gamma (1 - d_t) V(s_{t+1})\right),
\]
\[
L_{\text{actor}} = \mathbb{E}_{s_t \sim \mathcal{D}}\!\left[\alpha \log \pi(a_t | s_t) - Q(s_t, a_t)\right].
\]

\section{Temporal Attention Unit}
\label{appendix:TAU}

\subsection{Depthwise Convolution}
Depthwise convolution is a type of convolution that performs a single convolution per input channel, rather than mixing channels. This reduces computational cost, and architectures like MobileNet use it to build efficient models with strong performance. It is applied on an input tensor \( \mathbf{X} \in \mathbb{R}^{H \times W \times C}\) (with height \(H\), width \(W\), and channels \(C\)) as
\[
Y_c = X_c \times X_c, \quad \text{for } c = 1, \dots, C
\]
where \(K_c\) is the kernel corresponding to channel \(c\).

\subsection{Dilated Convolution}
After capturing channel-wise patterns via depthwise convolution, dilated convolution further expands the receptive field to capture long-range dependencies without increasing parameters. Dilated convolution introduces gaps between kernel elements and increases the receptive field without adding parameters. This enables the network to capture both short- and long-range dependencies as well as multi-scale context effectively. For a 1D input signal $x$ and a filter size $k$, a dilated convolution is defined as:
\[
(y *_d f)(i) = \sum_{j=0}^{k-1} f(j) \cdot x(i - d \cdot j).
\]
where \(d\) denotes the dilation rate.
\subsection{Pointwise Convolution}

Finally, pointwise 1$\times$1 convolution mixes information across channels and reduces dimensionality, enabling efficient feature integration. It helps mix information from different channels, reduce dimensionality, and make the model less complex; it also extracts more abstract features and is widely used in models like MobileNet that require efficiency and generalization. This convolution applies to an input tensor $X$ with $C$ channels, and a kernel $K \in \mathbb{R}^{1 \times 1 \times C \times C'}$; the pointwise convolution is defined as:
\[
Y(i, j, c') = \sum_{c=1}^{C} X(i, j, c) \cdot K(1, 1, c, c').
\]

\subsection{Multi-Head Self-Attention }
Multi-head self-attention, originally popularized in natural language processing, assigns varying importance to different elements of a sequence and is highly effective for time-series modeling. By capturing diverse relationships across temporal steps, it enhances prediction accuracy.
In multi-head self-attention, the model simultaneously focuses on different parts of the input, capturing diverse relationships and long-range dependencies.
This applies to an input $X$ and extracts queries $Q$, keys $K$, and values $V$ using linear transformations. Then, attention is defined as
\[
\text{Attention}(Q, K, V) = \text{softmax}\!\left(\frac{QK^T}{\sqrt{d_k}}\right)V
\]
where $d_k$ is the dimensionality of queries and keys. In multi-head attention, this process is executed in parallel multiple times, and results are concatenated to form the final output.

\end{document}